\documentclass[twocolumn,10pt]{asme2ej}

\usepackage{epsfig} 
\usepackage{lipsum}

\usepackage{epsfig}
\usepackage{amsmath}
\usepackage{amsfonts}
\usepackage{amssymb}
\usepackage{mathdots}
\usepackage{mathdots}
\usepackage[utf8]{inputenc}
\usepackage[titletoc,title]{appendix}
\usepackage{titletoc}

\usepackage{enumerate}
\usepackage{amsmath}
\usepackage{amssymb}
\usepackage{centernot}
\usepackage[mathscr]{euscript}
\usepackage{url}
\usepackage{hyphenat}
\usepackage{bm}
\usepackage{verbatim}
\usepackage{accents}
\usepackage{graphicx}
\usepackage{subcaption}
\usepackage{listings}
\usepackage{xcolor}
\usepackage{grffile}
\usepackage{rotating}
\usepackage{tikz}

\DeclareFontFamily{U}{mathx}{\hyphenchar\font45}
\DeclareFontShape{U}{mathx}{m}{n}{
      <5> <6> <7> <8> <9> <10>
      <10.95> <12> <14.4> <17.28> <20.74> <24.88>
      mathx10
      }{}
\DeclareSymbolFont{mathx}{U}{mathx}{m}{n}
\DeclareFontSubstitution{U}{mathx}{m}{n}
\DeclareMathAccent{\widecheck}{0}{mathx}{"71}
\DeclareMathAccent{\wideparen}{0}{mathx}{"75}

\usepackage{grffile}
\usepackage{algorithm,algorithmicx}
\usepackage{algpseudocode}
\algnewcommand\algorithmicinput{\textbf{Input: }}
\algnewcommand\algorithmicoutput{\textbf{Output: }}
\usepackage{afterpage}
\usepackage{epstopdf}
\usepackage{widetext}
\usepackage{cuted}
\usepackage{multicol}

\everymath{\displaystyle}

\DeclareMathOperator\erf{erf}

\newcommand\E[1]{{\mathbb{E}\left[#1\right]}}
\newcommand\V[1]{{\mathbb{V}\left[#1\right]}}

\newcommand\Cov[1]{{\text{Cov}\left[#1\right]}}

\usepackage{lineno,hyperref}
\modulolinenumbers[5]

\title{Monotonic Gaussian process for physics-constrained machine learning with materials science applications}

\author{Anh Tran\thanks{Corresponding author: Anh Tran (anhtran@sandia.gov)}
    \affiliation{
    Scientific Machine Learning \\
    Sandia National Laboratories \\
    Albuquerque, NM 87185 \\
    Email: anhtran@sandia.gov\\
    }
}

\author{Kathryn Maupin
    \affiliation{
    Optimization and Uncertainty Quantification \\
    Sandia National Laboratories \\
    Albuquerque, NM 87185 \\
    Email: kmaupin@sandia.gov\\
    }
}

\author{Theron Rodgers \\
    Computational Materials \& Data Science \\
    Sandia National Laboratories \\
    Albuquerque, NM 87185 \\
    Email: trodger@sandia.gov \\
}

\begin{document}

\maketitle
\makeatletter
\let\ps@oldempty\ps@empty 
\renewcommand\ps@empty\ps@plain
\makeatother


\begin{abstract}
Physics-constrained machine learning is emerging as an important topic in the field of machine learning for physics. One of the most significant advantages of incorporating physics constraints into machine learning methods is that the resulting model requires significantly less data to train. By incorporating physical rules into the machine learning formulation itself, the predictions are expected to be physically plausible. Gaussian process (GP) is perhaps one of the most common methods in machine learning for small datasets. In this paper, we investigate the possibility of constraining a GP formulation with monotonicity on three different material datasets, where one experimental and two computational datasets are used. The monotonic GP is compared against the regular GP, where significant reduction in the posterior variance is observed. The monotonic GP is strictly monotonic in the interpolation regime, but in the extrapolation regime, the monotonic effect starts fading away as one goes beyond the training dataset. Imposing monotonicity on the GP comes at a small accuracy cost, compared to the regular GP. The monotonic GP is perhaps most useful in applications where data is scarce and noisy, and monotonicity is supported by strong physical evidence.
\end{abstract}

\section{Introduction}

Physics-constrained machine learning is an important toolbox in the machine learning era, with many possible applications, including those in biology, materials science, astrophysics, and finance.
Breakthroughs in deep learning have been recognized as some of the most revolutionary contributions to science, but its Achilles heel is the size of the dataset required to train huge deep learning architectures.

However, this requirement cannot always be satisfied, especially in fields involving experimentation, where data can be resource-intensive, in terms of both time and labor.
In this context, physics-constrained machine learning arises as an alternative solution, with equivalent or better accuracy, while requiring significantly less data to train.
By incorporating some fundamental physics rules into the machine learning framework, the behavior of the machine learning predictions could be more physically plausible.
One common constraint is the monotonicity, where the predictions are expected to behave monotonically with respect to one or several input variables.

Materials science is a field where theory has overwhelmed in the last several decades and centuries.
The \textcolor{black}{process-structure-property} relationship is perhaps one of the most well-studied topics, in terms of theoretical analysis.
Existing well before the computer era, material scientists tended to agree that mathematics and physics are the right tools to bridge the gap between inputs and quantities of interest.
With the emergence of computers, supercomputers, and subsequently deep learning, the main tool has changed significantly, starting with the Materials Genome Initiative~\cite{national2011materials}.
However, brute-force applications of deep learning has suffered from the lack of data in materials science, where physics-constrained machine learning seems to be a more promising candidate.
Furthermore, materials science theory suggests that there are many simple physical rules that one can take advantage of when constructing a machine learning framework, for example, the Hall-Petch relationship~\cite{cordero2016six}.
It should be noted that many of these rules or simple equations are involve monotonicity.
If these rules are successfully exploited, then a physics-constrained machine learning could be achieved with much less data, while retaining the accuracy performance.

The Gaussian process (GP) stands out among other machine learning approaches and has been one of the most popular methods for small datasets across multiple disciplines.
In particular, it is well known that GPs are often used as underlying surrogate models for Bayesian optimization, which is an efficient active machine learning method and very popular among materials scientists.
For example, Tallman et al.~\cite{tallman2019gaussian,tallman2020uncertainty} used a GP as a surrogate model to bridge microstructure crystallography and homogenized materials properties, where crystal plasticity finite element models are used to construct the database, and
Yabansu et al.~\cite{yabansu2019application} applied GPs to capture the evolution of microstructure statistics.
Tran and Wildey~\cite{tran2020solving,tran2021solving1} also used a GP as a surrogate model for structure-property relationship and solved a stochastic inverse problem using the acceptance-rejection sampling method.
More examples can be found in~\cite{tran2020multi,khatamsaz2021efficiently}.

One of the main advantages of using a GP is its flexibility and, therefore, its ability to adopt extensions.
Fern{\'a}ndez-Godino et al.~\cite{fernandez2018use} proposed a novel approach to constrain symmetry on GPs.
Linear operator constraints can be enforced through a novel covariance function, as demonstrated by Jidling et al.~\cite{jidling2017linearly}.
Agrell~\cite{agrell2019gaussian} presented an approach to constrain GPs such that a set of linear transformations of the process are bounded.
Furthermore, Lange-Hegermann~\cite{lange2021linearly} constrained multi-output GP priors to the solution set of a system of linear differential equations subjected to boundary conditions.
A comprehensive survey study of constrained GPs, including, but not limited to, bound constraints, non-negativity, monotonicity, linear partial differential operator constraints, and boundary conditions of a partial differential equation can be found in Swiler, et al.~\cite{swiler2020survey}.

The present work focuses on bound-type constraints.
Riihim{\"a}ki and Vehtari~\cite{riihimaki2010gaussian} proposed enforcing monotonicity to constrain GPs using binary classification, where the derivative is classified as either monotonically increasing or decreasing with respect to a set of specific input variables.
Golchi et al.~\cite{golchi2015monotone} extended this approach in a deterministic computer experiments setting and sampled from the exact joint posterior distribution rather than an approximation.
More recently,
Ustyuzhaninov et al.~\cite{ustyuzhaninov2020monotonic} presented a monotonicity constraint in a Bayesian non-parametric setting based on numerical solutions of stochastic differential equations, and
Pensoneault et al.~\cite{pensoneault2020nonnegativity} presented a novel approach to construct a non-negative GP.
\textcolor{black}{Tan~\cite{tan2017monotonic} proposed a B-spline model that linearly constrains the model coefficients to achieve monotonicity, which converges to the projection of the GP in $L_2$.
Chen et al.~\cite{chen2021solving} proposed an optimal recovery method that is equivalent to maximum a posterior estimation for a GP constrained by a partial differential equation.
}

In this paper, we adopt the monotonic GP formulation from Riihim{\"a}ki and Vehtari~\cite{riihimaki2010gaussian} and apply it to three different datasets, both in interpolation and extrapolation regimes, to survey the advantages and disadvantages between the regular GP and the monotonic GP.
We focus on experimental and computational materials science applications where monotonicity is physically expected.
Materials science, specifically experimental materials science, is well-known for data scarcity due to its resource-intensive experiments.
An overview of Gaussian Processes and the incorporation of monotonic
constraints therein are discussed in Section~\ref{sec_gp_overview}. Three
examples of how classical and monotonic GPs compare in performance are given in
Sections~\ref{sec:ex:fatigue},~\ref{sec:ex:sppark}, and~\ref{sec:ex:damask}.
An overall discussion and conclusion are given in Section~\ref{sec:discussion}.






\section{Overview of Gaussian Processes}
\label{sec_gp_overview}

\subsection{Nomenclature}

The symbols used in this paper are as follows:

\begin{itemize}
\item $\mathbf{X} = \{\mathbf{x}^{(i)}\}_{i=1}^N$: training dataset of size $N$, $\mathbf{X} \in \mathbb{R}^{N \times D}$
\item $\mathbf{x} = [\mathbf{x}_1, \dots, \mathbf{x}_D]$: training input of $D$ dimensionality
\item $\mathbf{X}_m$: derivative inducing points, $\mathbf{X}_m \in \mathbb{R}^{m \times D}$
\item $\mathbf{y}$: noisy observation $\mathbf{y}(\mathbf{x}) = \mathbf{f}(\mathbf{x}) + \varepsilon$, $\varepsilon \sim \mathcal{N}(0,\sigma^2)$
\item $\mathbf{f}$: noiseless output $\mathbf{f}(\mathbf{x})$
\item $\mathbf{f}'$: first derivative of $\mathbf{f}$
\item $x^*$: testing input
\item $f^*$: testing output
\item $N$: number of data points
\item $M$: number of inducing data points for derivatives
\item $D$: input dimensionality
\item $i$, $j$: dummy data point index
\item $d$, $g$, $h$: dummy dimensionality index
\end{itemize}

\subsection{Classical Gaussian processes}

A thorough mathematical explanation and derivation of Gaussian processes can be found in the canonical text by Rasmussen and Williams~\cite{rasmussen2006gaussian}, but we provide a brief overview here.

Let us assume that we can model the true process, $\mathbf{y}$, with a zero-mean GP,
\begin{equation}
\mathbf{f} | \mathbf{\mathbf{X}} \sim \mathcal{N}(\mathbf{0}, \mathbf{K}_{\mathbf{f},\mathbf{f}}),
\end{equation}
where the entries in the covariance matrix $\mathbf{K}_{\mathbf{f}, \mathbf{f}}$ are defined by a
covariance function.
Although there are many covariance functions to choose from, in this paper, we focus on the squared exponential covariance function
\begin{equation}
\Cov{ f^{(i)} , f^{(j)} } = \mathbf{K}(\mathbf{x}^{(i)}, \mathbf{x}^{(j)}) = \eta^2 \exp\left[ - \frac{1}{2} \sum_{d=1}^D \rho_d^{-2} (x^{(i)}_d - x^{(j)}_d)^2 \right],
\label{eq:kernel}
\end{equation}
where $\eta$ and $\mathbf{\rho} = \{\rho_1, \dots, \rho_D \}$ are hyper-parameters, representing the signal variance and length-scale (also known as correlation length) parameters, respectively.
The observations $\mathbf{y}$ are then given by
\begin{equation}
\mathbf{y} | \mathbf{f} \sim \mathcal{N}(\mathbf{f}, \sigma^2 \mathbf{I}),
\end{equation}
where $\sigma^2$ is the intrinsic noise variance.

At a desired input prediction location, $x^{*}$, the posterior prediction distribution, also called the testing distribution, is a Gaussian distribution, where the posterior mean and posterior variance are, respectively,
\begin{equation}
\E{f^* | x^*, \mathbf{y}, \mathbf{X}, \mathbf{\theta}} = \mathbf{K}_{\mathbf{*},\mathbf{f}} [\mathbf{K}_{\mathbf{f},\mathbf{f}} + \sigma^2 \mathbf{I}]^{-1} \mathbf{y}
\end{equation}
and
\begin{equation}
\V{f^* | x^*, \mathbf{y}, \mathbf{X}, \mathbf{\theta}} = \mathbf{K}_{\mathbf{*},\mathbf{*}} - \mathbf{K}_{*,\mathbf{f}} [\mathbf{K}_{\mathbf{f},\mathbf{f}} + \sigma^2 \mathbf{I}]^{-1} \mathbf{K}_{\mathbf{f},\mathbf{*}}.
\end{equation}
Here, the hyper-parameters are collected into $\mathbf{\theta} = \{ \eta, \mathbf{\rho}, \sigma\}$.

The GP is trained by solving for the hyper-parameters $\mathbf{\theta}$ that maximize the log-marginal likelihood
\begin{equation}
\begin{array}{lll}
\log p(\mathbf{y} | \mathbf{X}, \mathbf{\theta}) &=& - \frac{1}{2}\mathbf{y}^\top [\mathbf{K}_{\mathbf{f},\mathbf{f}} + \sigma^2 \mathbf{I}]^{-1} \mathbf{y} \\
&& - \frac{1}{2} \log{|\mathbf{K}_{\mathbf{f}, \mathbf{f}} + \sigma^2 \mathbf{I}|} - \frac{N}{2} \log{(2\pi)}
\end{array}
\label{eq:mle}
\end{equation}
In Equation~\ref{eq:mle}, the first term, so-called the ``data fit" term, quantifies how well the model fits the data described in the Mahalanobis distance. The second term, called the ``complexity" term, quantifies the model complexity where smoother covariance matrix with smaller determinant is encouraged~\cite{rasmussen2006gaussian}. The last term indicates that the likelihood of data tends to decrease with larger data sets~\cite{shahriari2016taking}.

It is noteworthy that the derivative operator is a linear operator; therefore,
\begin{equation}
\E{ \frac{\partial f^{(i)}}{\partial x_d^{(i)}} } =  \frac{\partial \E{f^{(i)}}}{\partial x_d^{(i)}},
\end{equation}
\begin{equation}
\Cov{ \frac{\partial f^{(i)}}{\partial x_d^{(i)}} , f^{(j)} } =  \frac{\partial }{\partial x_d^{(i)}} \Cov{f^{(i)}, f^{(j)}},
\label{eq:covFirstDerivative}
\end{equation}
and
\begin{equation}
\Cov{ \frac{\partial f^{(i)}}{\partial x_d^{(i)}} , \frac{\partial f^{(j)}}{\partial x_g^{(j)}} } =  \frac{\partial^2 }{\partial x_d^{(i)} \partial x_g^{(j)} } \Cov{f^{(i)}, f^{(j)}}.
\label{eq:covSecondDerivative}
\end{equation}
For the squared exponential kernel described in Equation~\ref{eq:kernel}, Equations~\ref{eq:covFirstDerivative} and~\ref{eq:covSecondDerivative} become
\begin{equation}
\begin{array}{lll}
\Cov{ \frac{\partial f^{(i)}}{\partial x_g^{(i)}} , f^{(j)} } &=& - \eta^2 \exp\left( -\frac{1}{2} \sum_{d=1}^D \rho_d^{-2} (x_d^{(i)} - x_d^{(j)})^2 \right) \\
&& \rho_g^{-2} \left( (x_g^{(i)} - x_g^{(j)}) \right) ,
\end{array}
\end{equation}
and
\begin{equation}
\begin{array}{lll}
\Cov{ \frac{\partial f^{(i)}}{\partial x_d^{(i)}} , \frac{\partial f^{(j)}}{\partial x_h^{(j)}} } &=& \eta^2 \exp\left( -\frac{1}{2} \sum_{d=1}^D \rho_d^{-2} (x_d^{(i)} - x_d^{(j)})^2 \right) \\
&& \rho_g^{-2} \left( \delta_{gh} - \rho_h^{-2} (x_g^{(i)} - x_g^{(j)}) (x_h^{(i)} - x_h^{(j)}) \right) ,
\end{array}
\end{equation}
respectively, where $\delta_{gh} = 1$ if $g=h$ and 0 otherwise.

For an arbitrary testing point $x^*$, the derivatives with respect to the dimension of the posterior mean and posterior variance are, respectively,
\begin{equation}
\E{ \frac{\partial f^*}{\partial x^*_d} } = \frac{\partial \mathbf{K}_{*, \mathbf{f}}}{\partial x^*_d} [\mathbf{K}_{\mathbf{f}, \mathbf{f}} + \sigma^2 \mathbf{I}]^{-1} \mathbf{y},
\end{equation}
and
\begin{equation}
\begin{array}{lll}
\V{ \frac{\partial f^*}{\partial x^*_d} } =  \frac{\partial^2 \mathbf{K}_{*, *}}{\partial x^*_d \partial x^*_d} - \frac{\partial \mathbf{K}_{*, \mathbf{f}}}{\partial x^*_d} [\mathbf{K}_{\mathbf{f}, \mathbf{f}} + \sigma^2 \mathbf{I}]^{-1} \frac{\partial \mathbf{K}_{\mathbf{f}, *}}{\partial x^*_d}.
\end{array}
\end{equation}

\subsection{Monotonic Gaussian processes}
\label{subsec:monotonicGP}

In this work, we adopt the monotonic GP formulation from Riihim{\"a}ki and Vehtari~\cite{riihimaki2010gaussian}, which has been implemented in the GPstuff toolbox~\cite{vanhatalo2012bayesian,vanhatalo2013gpstuff}.
To constrain the classical GP to be monotonic, Riihim{\"a}ki and Vehtari proposed a block covariance matrix structure similar to that of many multi-fidelity approaches, e.g. ~\cite{tran2020smfbo2cogp,yang2020bifidelity,xiao2018extended}, and approximated the posterior using expectation propagation~\cite{minka2001expectation}, which is arguably more efficient than Laplace's method. For the sake of completeness, we assume a zero-mean GP and briefly summarize the formulation. Readers are referred to the original work of Riihim{\"a}ki and Vehtari~\cite{riihimaki2010gaussian} for further details.
\textcolor{black}{Roughly speaking, the crux of the approach is to augment the covariance matrix $\mathbf{K}$ using a block structure, as is done in multi-fidelity and gradient-enhanced GP. Now, the difference is that the augmented block encodes the binary classification of whether the GP is supposed to be monotonically increasing or decreasing at some locations. Naturally, the formulation of GP lends itself into binary classification problem with the linear logistic regression or the probit regression, taken from the cumulative density function of a standard normal distribution $\Phi(z) = \int_{-\infty}^t \mathcal{N}(t | 0,1) dt$.
}

Monotonicity is imposed at $M$ inducing locations $\mathbf{X}_m \in \mathbb{R}^{M \times D}$. At the location $\mathbf{x}^{(i)} \in \mathbf{X}_m$, the derivative of $\mathbf{f}$ is non-negative with respect to the input dimension $d_i$. The key idea is to assume a probit likelihood at the location $\mathbf{x}^{(i)}$ as
\begin{equation}
p\left( m_{d_i}^{(i)} \Bigg| \frac{\partial f^{(i)}}{\partial x_{d_i}^{(i)}} \right) = \Phi\left( \frac{\partial f^{(i)}}{\partial x_{d_i}^{(i)}} \frac{1}{\nu} \right),
\label{eq:probitMonotonicity}
\end{equation}
where
\begin{equation}
\Phi(z) = \frac{1}{2} \left[ 1 + \erf\left( \frac{z}{\sqrt{2}} \right) \right] = \int_{-\infty}^{z} \mathcal{N}(t | 0,1) dt
\end{equation}
is the cumulative distribution function of the standard normal distribution.
Conceptually, imposing the probit likelihood is intrinsically similar to classification using GP (cf. Section 3.6~\cite{rasmussen2006gaussian} and Kuss et al.~\cite{kuss2005assessing} for binary classification \{-1,+1\} using GP).
Put simply, Equation~\ref{eq:probitMonotonicity} models the probability of ``penalized" monotonicity with the parameter $\nu$, in the same way one would in binary classification~\cite{kuss2005assessing}, where $\nu = 10^{-6}$ is recommended.

Assume that at $\mathbf{X}_m$, the function $\mathbf{f}$ is known to be monotonic. The joint prior for $\mathbf{f}$ and its derivatives $\mathbf{f}'$ is given by
\begin{equation}
p(\mathbf{f}, \mathbf{f}' | \mathbf{X}, \mathbf{X}_m) = \mathcal{N} \left(
\mathbf{f}_{\text{joint}} |  \mathbf{0}, \mathbf{K}_{\text{joint}}
\right),
\end{equation}
where
\begin{equation}
\mathbf{f}_{\text{joint}} = \begin{bmatrix} \mathbf{f} \\ \mathbf{f}' \end{bmatrix},
\quad
\mathbf{K}_{\text{joint}} = \begin{bmatrix} \mathbf{K}_{\mathbf{f}, \mathbf{f}} & \mathbf{K}_{\mathbf{f}, \mathbf{f}'} \\ \mathbf{K}_{\mathbf{f}', \mathbf{f}} & \mathbf{K}_{\mathbf{f}', \mathbf{f}'} \end{bmatrix}.
\end{equation}
Using the Bayes rule, the joint posterior is
\begin{equation}
p(\mathbf{f}, \mathbf{f}' | \mathbf{y}, \mathbf{m}) = \frac{1}{Z} p(\mathbf{f}, \mathbf{f}' | \mathbf{X}, \mathbf{X}_m) p(\mathbf{y} | \mathbf{f}) p(\mathbf{m} | \mathbf{f}'),
\label{eq:truePosterior}
\end{equation}
where
\begin{equation}
p(\mathbf{m} | \mathbf{f}') = \prod_{i=1}^M \Phi\left( \frac{\partial f^{(i)}}{\partial x_{d_i}^{(i)}} \frac{1}{\nu} \right).
\end{equation}
Since the posterior is analytically intractable, local likelihood approximations are given by the expectation propagation (EP) algorithm, allowing the approximation of the posterior distribution in Equation~\ref{eq:truePosterior}
\begin{equation}
\begin{array}{lll}
p(\mathbf{f}, \mathbf{f}' | \mathbf{y}, \mathbf{m}) &\approx& q(\mathbf{f}, \mathbf{f}' | \mathbf{y}, \mathbf{m}) \\
&=& \frac{1}{Z_{\text{EP}}} p(\mathbf{f}, \mathbf{f}' | \mathbf{X}, \mathbf{X}_m) p(\mathbf{y} | \mathbf{f}) \prod_{i=1}^M t_i(\tilde{Z}_i, \tilde{\mu}_i, \tilde{\sigma}^2_i),
\end{array}
\label{eq:approxPosterior}
\end{equation}
where $t_i(\tilde{Z}_i, \tilde{\mu}_i, \tilde{\sigma}^2_i) = \tilde{Z}_i \mathcal{N}(f'_i | \tilde{\mu}_i, \tilde{\sigma}^2_i)$ are local likelihood approximations with site parameters $\tilde{Z}_i, \tilde{\mu}_i, \tilde{\sigma}$ from the EP algorithm.

The approximate posterior is analytically tractable as a product of Gaussian distributions and can be simplified to
\begin{equation}
q(\mathbf{f}, \mathbf{f}' | \mathbf{y}, \mathbf{m}) = \mathcal{N}(\mathbf{f}_\text{joint} | \mathbf{\mu}, \Sigma),
\end{equation}
where
\begin{equation}
\mathbf{\mu} = \Sigma \tilde{\Sigma}_{\text{joint}}^{-1} \tilde{\mathbf{\mu}}_{\text{joint}},
\quad
\Sigma = [\mathbf{K}_{\text{joint}}^{-1} + \tilde{\Sigma}_{\text{joint}}^{-1}]^{-1},
\end{equation}
and
\begin{equation}
\tilde{\mathbf{\mu}}_\text{joint} = \begin{bmatrix} \mathbf{y} \\ \tilde{\mathbf{\mu}} \end{bmatrix},
\quad
\tilde{\Sigma}_\text{joint} = \begin{bmatrix} \sigma^2 \mathbf{I} & 0 \\ 0 & \tilde{\Sigma} \end{bmatrix},
\end{equation}
where $\tilde{\mu}$ is the vector of site means $\tilde{\mu}_i$, and $\tilde{\Sigma} = \text{Diag}[\tilde{\sigma}_i^2]_{i=1}^M$.
The approximation for the logarithm of the marginal likelihood is computed as
\begin{equation}
\begin{array}{lll}
\log Z_\text{EP} &=& -\frac{1}{2} \log |\mathbf{K}_\text{joint} + \tilde{\Sigma}_\text{joint}| \\
&& - \frac{1}{2} \tilde{\mathbf{\mu}}_\text{joint}^\top [\mathbf{K}_\text{joint} + \tilde{\Sigma}_\text{joint}]^{-1} \tilde{\mathbf{\mu}}_\text{joint} \\
&& + \sum_{i=1}^M \frac{(\mu_{-i} - \tilde{\mu}_i)^2}{2 (\sigma^2_{-i} + \tilde{\sigma}_i^2)} \\
&& + \sum_{i=1}^M \log \Phi\left( \frac{\mu_{-i}}{\nu \sqrt{1 + \sigma^2_{-i}/\nu^2}} \right) \\
&& + \frac{1}{2} \sum_{i=1}^M \log(\sigma^2_{-i} + \tilde{\sigma}^2_i),
\end{array}
\end{equation}
where $\mu_{-i}$ and $\sigma^2_{-i}$ are the parameters of the cavity distribution in EP.
By introducing $M$ monotonic inducing point, the cost complexity for optimizing the log-marginal likelihood increases from $\mathcal{O}\left(N^3\right)$ to $\mathcal{O}\left((N+M)^3\right)$.
The posterior mean and posterior variance of the testing distribution are, respectively, given by
\begin{equation}
\E{f^* | x^*, \mathbf{y}, \mathbf{X}, \mathbf{m}, \mathbf{X}_m} = \mathbf{K}_{*,\text{joint}} [\mathbf{K}_{\text{joint}} + \tilde{\Sigma}_\text{joint}]^{-1} \tilde{\mu}_{\text{joint}}
\label{eq:posteriorMeanMonotonic}
\end{equation}
and
\begin{equation}
\V{f^* | x^*, \mathbf{y}, \mathbf{X}, \mathbf{m}, \mathbf{X}_m} = \mathbf{K}_{*,*} - \mathbf{K}_{*,\text{joint}} [\mathbf{K}_{\text{joint}} + \tilde{\Sigma}_\text{joint}]^{-1} \mathbf{K}_{*,\text{joint}}.
\label{eq:posteriorVarMonotonic}
\end{equation}

\section{Numerical examples}

\subsection{Example 1: Noisy logistic regression}

\begin{figure}[!htbp]
\centering
\includegraphics[width=0.5\textwidth]{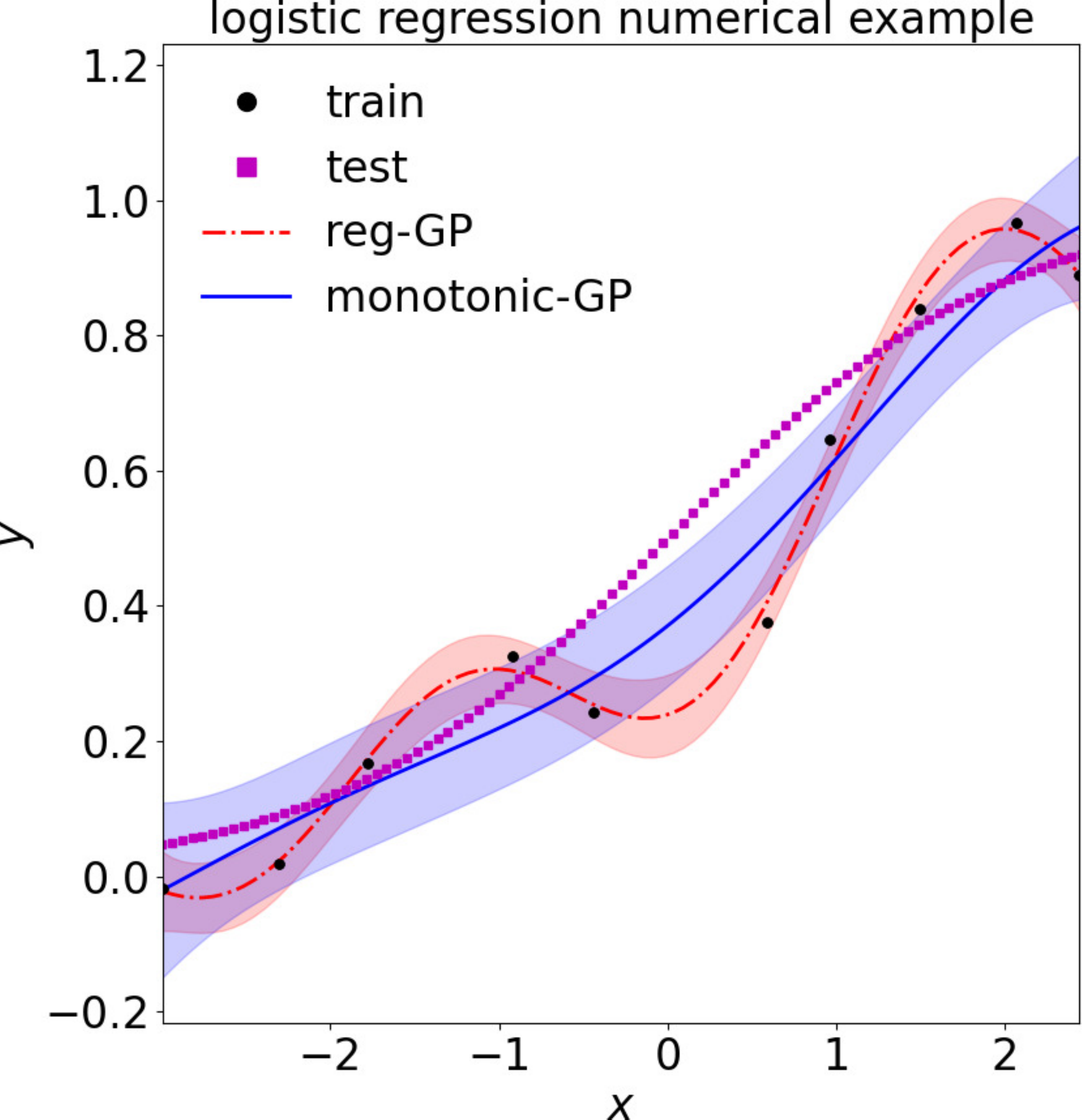}
\caption{\textcolor{black}{Comparison between the regular and monotonic GPs in the homoscedastic noisy logistic regression example. Training and testing data points are shown as black dots and magenta squares, respectively. By enforcing the monotonicity, the monotonic GP has a better approximation compared to the regular GP.}}
\label{fig:cropped_logistic}
\end{figure}

\textcolor{black}{
In this example, we consider the 1d monotonic function
\begin{equation}
y = \frac{1}{1 + e^{-x}} + \varepsilon,
\end{equation}
where $\varepsilon \sim \mathcal{N}(0,0.1^2) $ on [-3,3] with 10 samples, as shown in Figure~\ref{fig:cropped_logistic}.
Figure~\ref{fig:cropped_logistic} shows the comparison between the regular (dashed red) and monotonic GP (solid blue), where the training dataset is plotted as black dots, and testing dataset is plotted as magenta squares.
The noise $\varepsilon$ is substantial enough to corrupt the monotonicity of the underlying function.
Despite the fact that the noise is homoscedastic, i.e. the variance of the noise is constant throughout the bounded domain, the regular GP has not been able to capture the function correctly.
On the other hand, the monotonic GP enforces the monotonicity while ignoring the noise, which results in a better approximation of the true function, which is also plotted as magenta squares as testing dataset.
}

\subsection{Example 2: Heteroscedastic Hall-Petch relationship}

\begin{figure}[!htbp]
\centering
\includegraphics[width=0.5\textwidth]{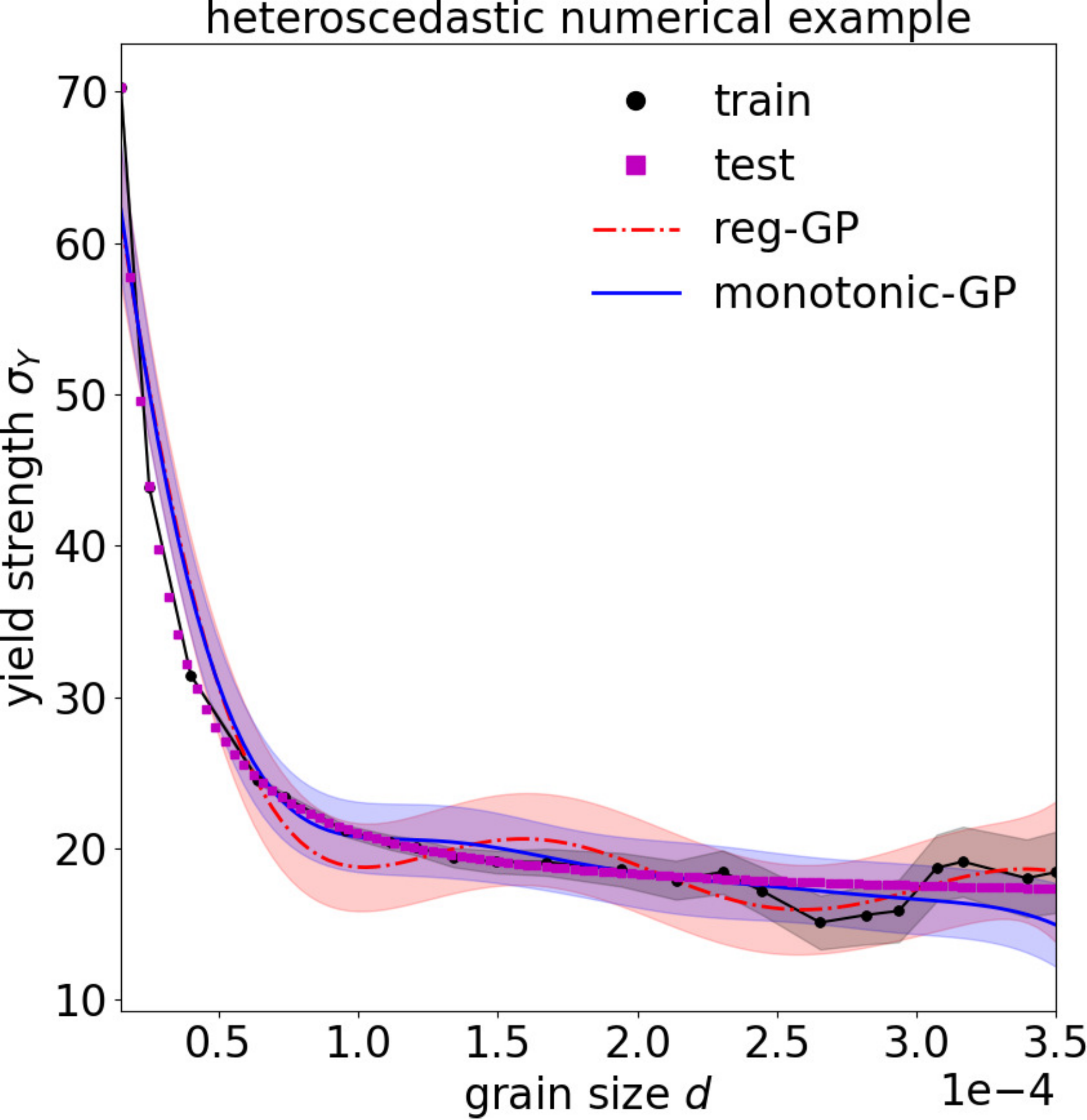}
\caption{\textcolor{black}{Comparison between the regular and monotonic GPs in the heteroscedastic settings. The noise $\varepsilon(d)$ increases as $d$ increases. It is observed that the regular GP is strongly influence.}}
\label{fig:cropped_heteroscedastic_HallPetch}
\end{figure}

\textcolor{black}{
Following Counts et al. ~\cite{counts2008predicting}, we adopt the Hall-Petch relationship in polycrystalline materials from Fernandes and Vieira~\cite{fernandes2000further} with a heteroscedastic noise proportional to $d$,
\begin{equation}
\sigma_Y = 16.47 + 0.0000288 \frac{1}{(10^{-6} \cdot d)^{1.3}} + \varepsilon(d).
\end{equation}
where $\varepsilon(d) \sim \mathcal{N}(0,  2.2 \cdot 10^{10} \cdot d^{3})$, to model the finite-size effect of the representative volume element~\cite{tran2020solving}.
$d$ is considered on the [15$\mu$ m, 350$\mu$ m].
For a fixed-size representative volume element, when $d$ is large, the number of grain reduces, and therefore, the noise of $\sigma_Y$ increases.
The same argument applies when $d$ is small.
Now, the variance of the Monte Carlo estimator roughly decays at the rate of $N^{-1}$ where $N$ is the number of grains or samples; in this hypothetical example, to examine the behavior of the monotonic GP, we simply model the volume of the grain as proportional to $d^3$ and therefore, the noise term with zero-mean and $\mathcal{O}(d^{3})$ variance is considered.
Obviously, the variance of the noise term increases as $d$ increases, which is consistent to what we have observed in the literature~\cite{tran2020solving}.
}

\textcolor{black}{
Figure~\ref{fig:cropped_heteroscedastic_HallPetch} shows the comparison between the regular and monotonic GP with 20 training data points, plotted as black dots.
The regular GP overfits the training dataset, which results in an oscillatory behavior toward the large $d$ region, which is unphysical.
The monotonic GP, on the other hand, correctly identifies the trend and monotonically decreases as $d$ increases, which results in a better approximation compared to the regular GP.
}

\section{Case study: Fatigue life prediction under multiaxial loading}
\label{sec:ex:fatigue}

\subsection{Dataset}

\begin{table}
\centering
\begin{tabular}{|l|l|l|l|} \hline
\multicolumn{2}{|l|}{Training} & \multicolumn{2}{|l|}{Testing} \\ \hline
$\sigma_a$ [MPa]  &  $N_{\text{exp}}$  &  $\sigma_a$ [MPa]  &  $N_{\text{exp}}$    \\ \hline \hline
674 & 2908 & 427 & 77730 \\
558 & 8115 & 400 & 113900 \\
556 & 10035 & 411 & 117275 \\
507 & 17012 & 403 & 144264 \\
483 & 19955 & 390 & 192920 \\
505 & 20595 & 391 & 198992 \\
498 & 23780 & 379 & 243816 \\
490 & 25913 & 366 & 376815 \\
484 & 28045 & 369 & 396987 \\
474 & 51430 & 345 & 406800 \\
469 & 52000 & 342 & 1252208 \\
475 & 66200 & 335 & 1444998 \\
    &       & 335 & 1528487 \\ \hline
\end{tabular}
\caption{Training and testing fatigue datasets for S355N steel, adopted from Karolczuk and S{\l}o{\'n}ski~\cite{karolczuk2022application}.}
\label{tab:fatigueS355Ndataset}
\end{table}

We adopt a subset from Karolczuk and S{\l}o{\'n}ski~\cite{karolczuk2022application,karolczuk2020application}, where the training and testing datasets are listed in Table~\ref{tab:fatigueS355Ndataset}.
The training dataset is taken to be 12 data points with high $\sigma_a$, whereas 13 data points with low $\sigma$ are used as the testing dataset.
We are interested in investigating the extrapolation capability of GPs beyond their training regime as well as their robustness against noise, which is fairly common in experimental materials science, particularly in fatigue, where stochasticity plays a non-trivial role.
Much of the stochasticity can be traced back to the physics of fracture and fatigue, where void nucleation, void growth, and void coalescence all play important roles in the origin, development, and failure of a material.
Due to this underlying stochasticity, sometimes the data shows a substantially noisy behavior despite a clear physical monotonic trend.
That is, the higher the stress amplitude is, the lower the fatigue life is.
This behavior is typically modeled by the SN equation as
\begin{equation}
\log N_{\text{exp}} = A - B \log \sigma_a,
\end{equation}
where $A$ and $B$ are material-dependent coefficients, $\sigma_a$ is the stress amplitude, and $N_{\text{f}}$ is the experimental fatigue life.
Experimental materials science is resource-intensive~\cite{arroyave2019systems}; thus, practically speaking, fatigue experimental data is scarce and may not be strictly monotonic, as shown in Karolczuk and S{\l}o{\'n}ski~\cite{karolczuk2022application}.
In the present work, the input of the GPs is $\sigma_a$, and the output of the GPs is $\log N_\text{exp}$.

\subsection{Results}

\begin{figure}
\centering
\includegraphics[width=0.5\textwidth]{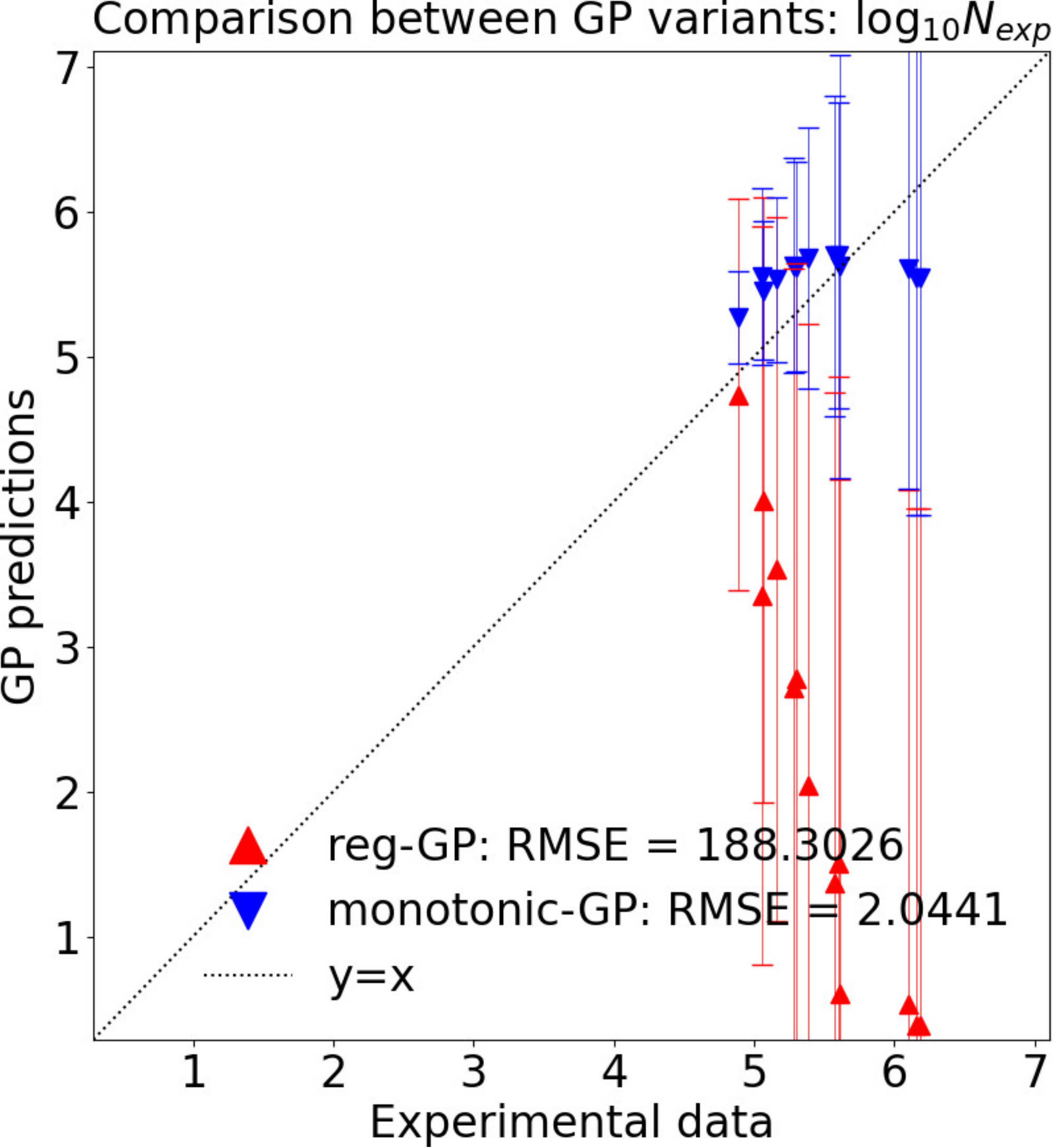}
\caption{Accuracy comparison of testing dataset predictions between the regular and monotonic GPs in S355N fatigue dataset, where the monotonic GP outperforms the regular GP.}
\label{fig:cropped_gp_s355n_comparison}
\end{figure}

\begin{figure}
\centering
\includegraphics[width=0.5\textwidth]{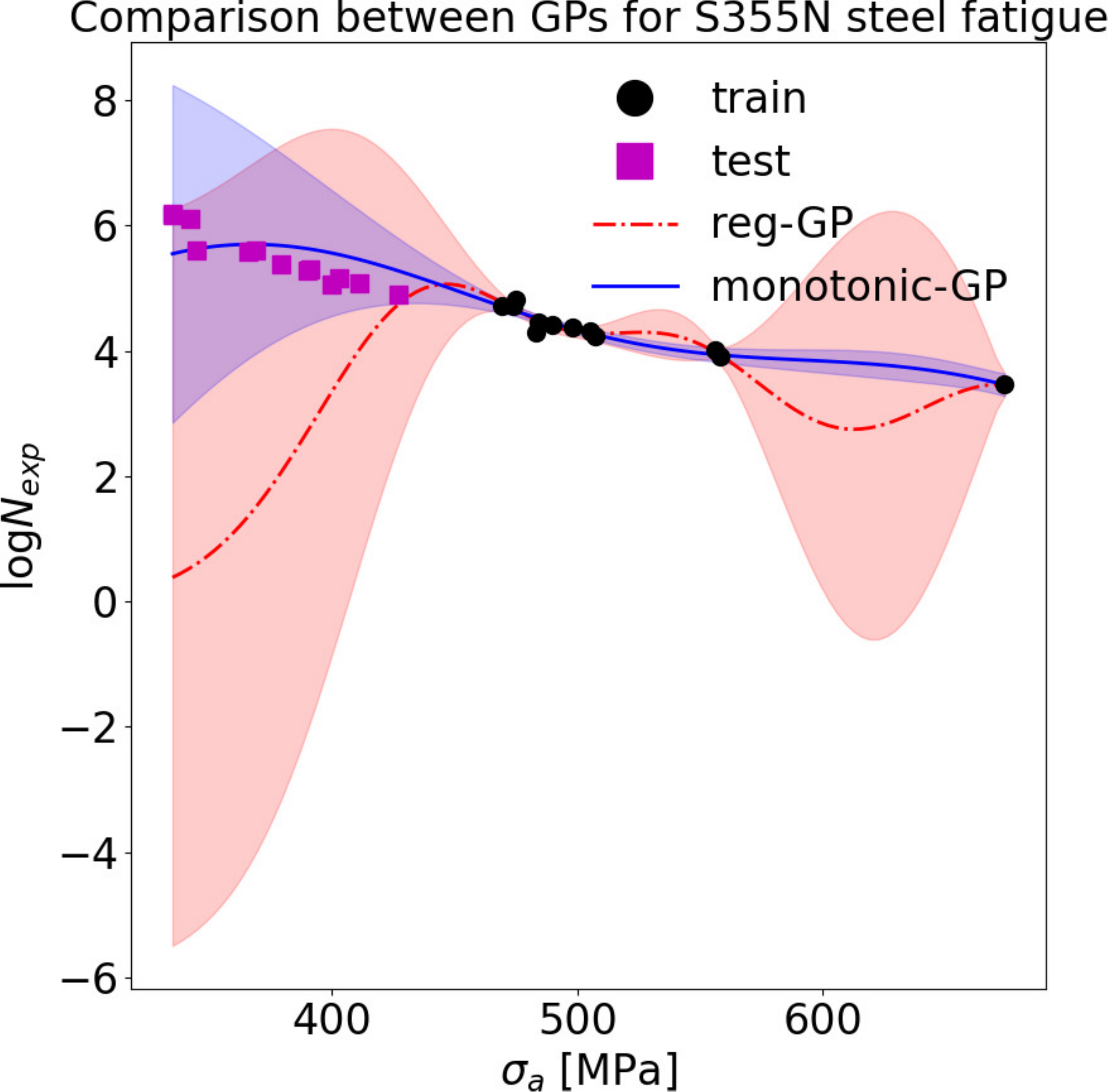}
\caption{Comparison between the regular and monotonic GPs in S355N fatigue dataset. $\mu \pm 1\sigma$ confidence intervals are highlighted. It is observed that the monotonicity starts to disappear as the extrapolation range goes further to $\sigma_a \leq 375$MPa.}
\label{fig:cropped_gp_input_output_fatigue_comparison}
\end{figure}

We implement and compare two GP variants: the regular GP and a monotonically-constrained GP. Both are trained using a squared exponential kernel and the hyper-parameters are optimized using the L-BFGS method~\cite{zhu1997algorithm} with a maximum of 6000 iterations.
Figure~\ref{fig:cropped_gp_s355n_comparison} compares the accuracy between the two GPs against the testing dataset.
The root-mean-square errors for the regular GP and the monotonic GP are 183.3026 and 2.0441, respectively.
While the regular GP tends to underestimate the experimental data, the monotonic GP provides more accurate predictions.
As shown in Figure~\ref{fig:cropped_gp_input_output_fatigue_comparison}, outside the training domain, the regular GP is unable to capture the general trend and behaves wildly, even when interpolating the data points.
On the contrary, the monotonic GP is able to capture the monotonic trend and extrapolate to a significant distance outside the training interval.
It is important to note that the monotonic GP is only strictly monotonic within the training domain and may fail to maintain monotonicity far beyond the training regime.
Indeed, in Figure~\ref{fig:cropped_gp_input_output_fatigue_comparison}, we see that the monotonic GP fails to maintain monotonicity when $\sigma_a \lesssim 350$ MPa.
Figure~\ref{fig:cropped_gp_input_output_fatigue_comparison} also shows a non-monotonic behavior for the regular GP.

It is also observed that the posterior variance of the monotonic GP is significantly less than the posterior variance of the classic GP, using the same training dataset.
This holds true for both interpolation and extrapolation cases.
The main reason is that the monotonic GP uses an extra ``pseudo'' training dataset by imposing $M$ inducing location in the input domain, as described in Section~\ref{subsec:monotonicGP}, and a training dataset with more data points reduces the posterior variance.

\section{Case study: Potts Kinetic Monte Carlo for grain growth}
\label{sec:ex:sppark}

\subsection{Model description}

The details of the Potts kinetic Monte Carlo (KMC) simulation for grain growth and its implementation in SPPARKS is described in \cite{garcia2008three,plimpton2009crossing}, and is summarized here for the sake of completeness.
In the grain growth simulation, the Potts model \cite{anderson1989computer} is used to simulate curvature-driven grain growth. Grain microstructures are represented by an integer value stored at each pixel. During a timestep (referred to here as a Monte Carlo or MC step), pixels in the simulation are visited and probabilistically change their grain membership to a neighboring grain based on the Metropolis algorithm.
The probability $P$ of successful change in grain site orientation is calculated as
\begin{equation}
P =
\begin{cases}
\exp\left( \frac{-\Delta E}{k_B T_s} \right), & \text{ if } \Delta E > 0, \\
1, &  \text{ if } \Delta E \leq 0,
\end{cases}
\label{eq:Metropolis}
\end{equation}
where $E$ is the total grain boundary energy calculated by summing all the neighbor interaction energies, $\Delta E$ can be regarded as the activation energy, $k_B$ is the Boltzmann constant, and $T_s$ is the simulation temperature. In the basic Potts model, the interaction energy between two pixels belonging to the same grain is zero, and $E$ is incremented by one for each dissimilar neighbor. From Equation \ref{eq:Metropolis}, changes that decrease system energy are preferred, and the total system energy is monotonically decreased through grain coarsening.
It is worthy to note that the $T_s$ simulation temperature is not the real system temperature: $k_B T_s$ is an energy that defines the thermal fluctuation, i.e. noise, presented in the kMC simulation~\cite{garcia2008three}. Increasing the value of $k_B T_s$ results in microstructures with increasing grain boundary roughness.

\begin{figure}

\centering
\subcaptionbox{SPPARKS at 50 MC step.
\label{fig:spparks50mcs}
}
  [0.495\linewidth]{\includegraphics[width=0.24\textwidth,keepaspectratio]{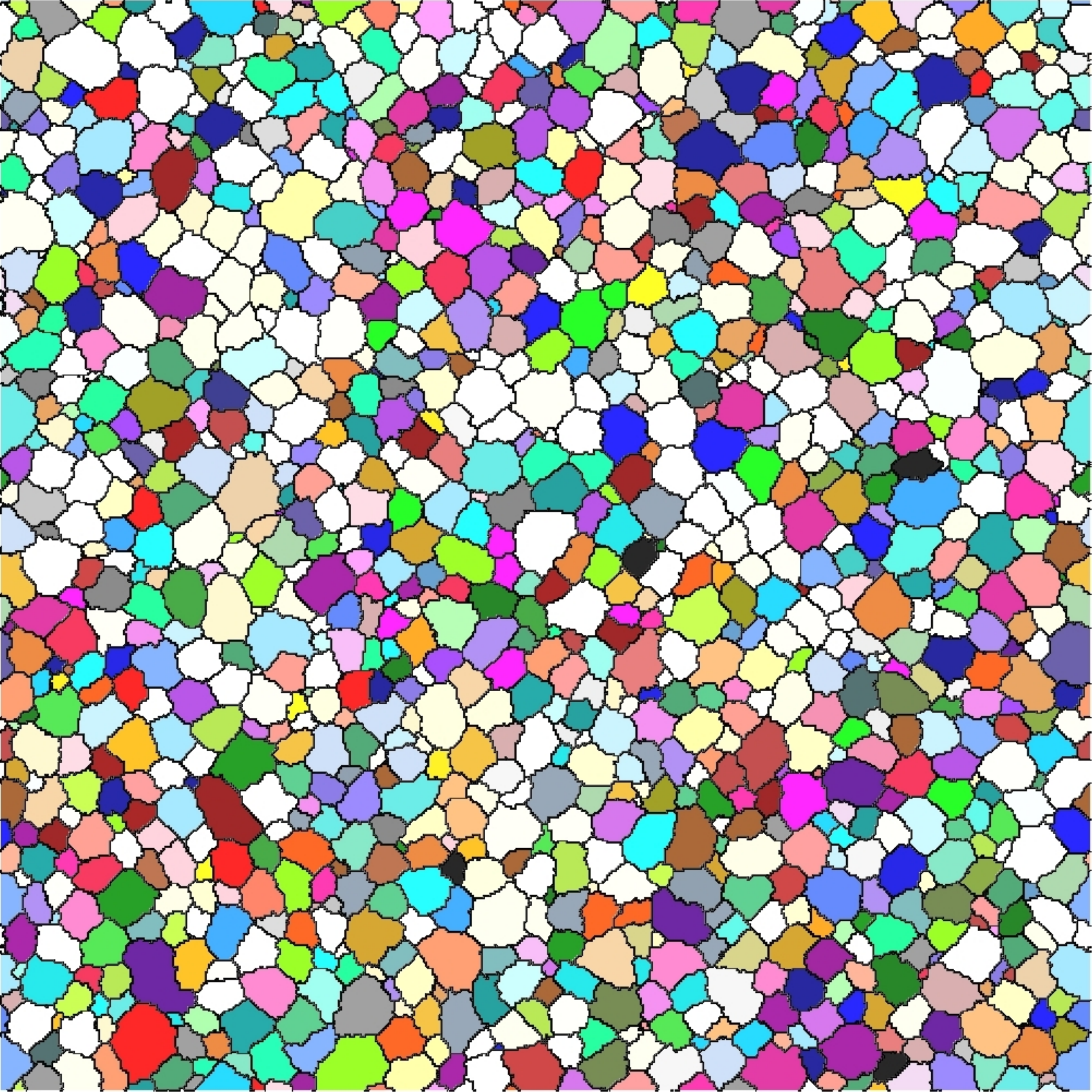}}
\hfill
\subcaptionbox{SPPARKS at 100 MC step.
\label{fig:spparks100mcs}
}
  [0.495\linewidth]{\includegraphics[width=0.24\textwidth, keepaspectratio]{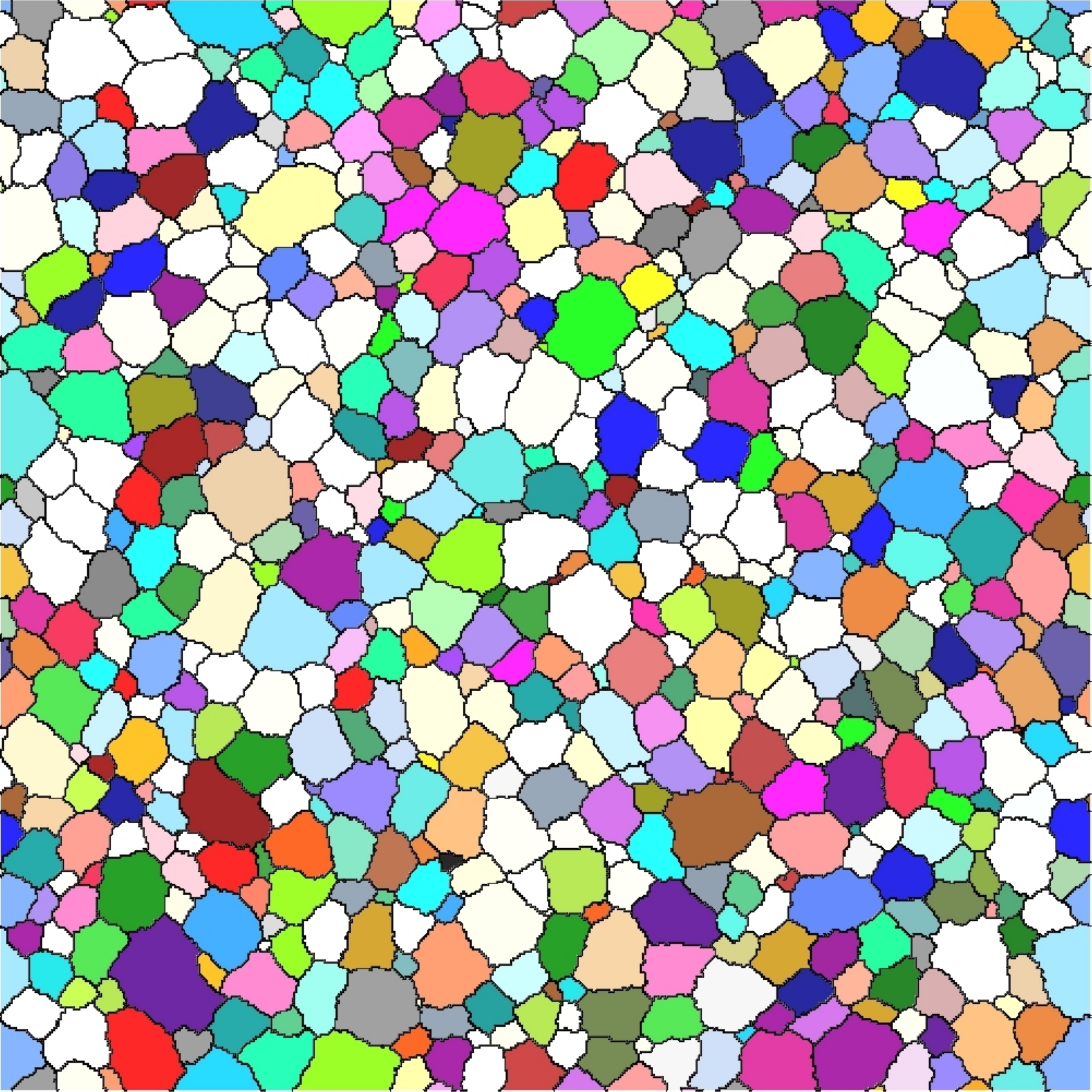}}
\vfill
\subcaptionbox{SPPARKS at 200 MC step.
\label{fig:spparks200mcs}
}
  [0.495\linewidth]{\includegraphics[width=0.24\textwidth,keepaspectratio]{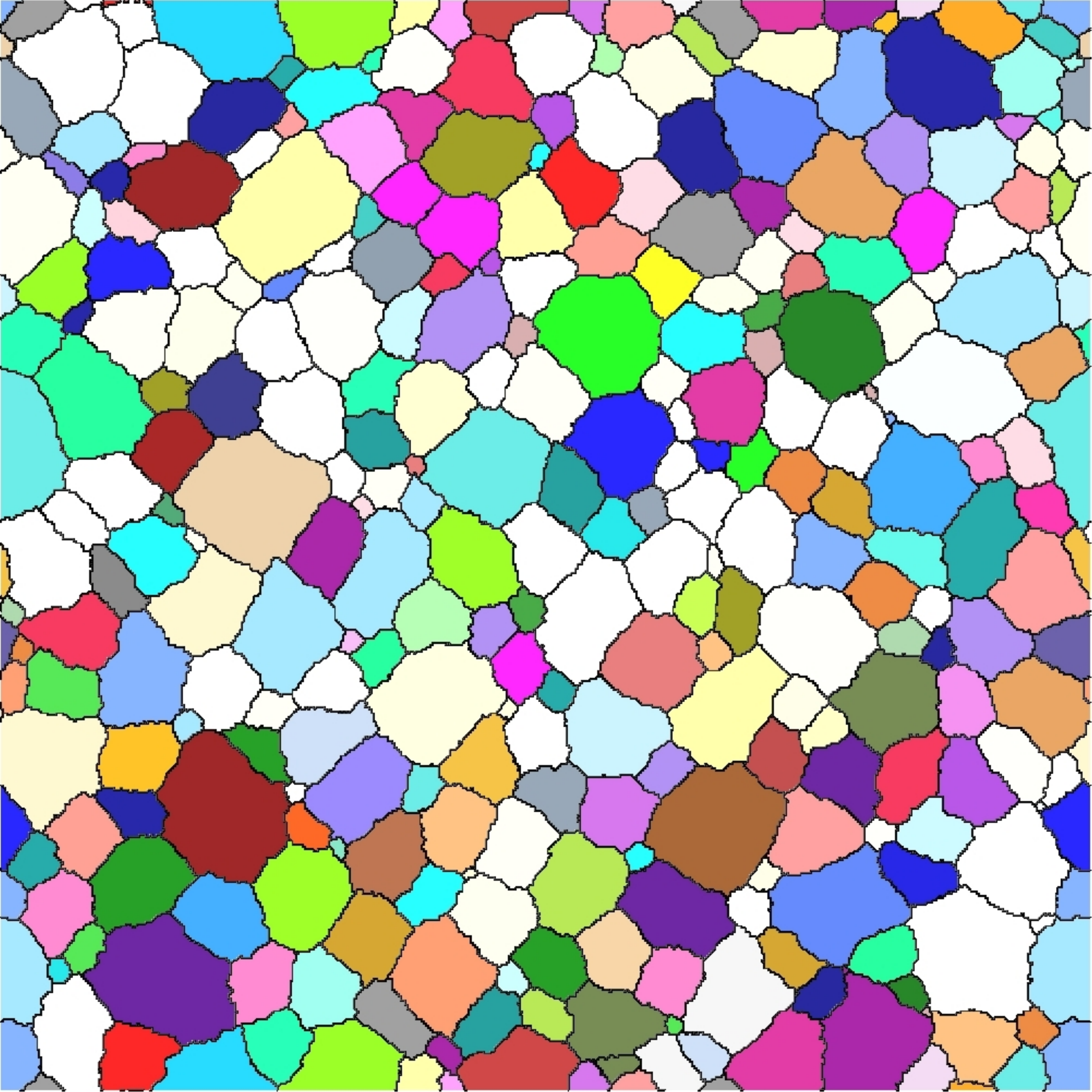}}
\hfill
\subcaptionbox{SPPARKS at 350 MC step.
\label{fig:spparks300mcs}
}
  [0.495\linewidth]{\includegraphics[width=0.24\textwidth, keepaspectratio]{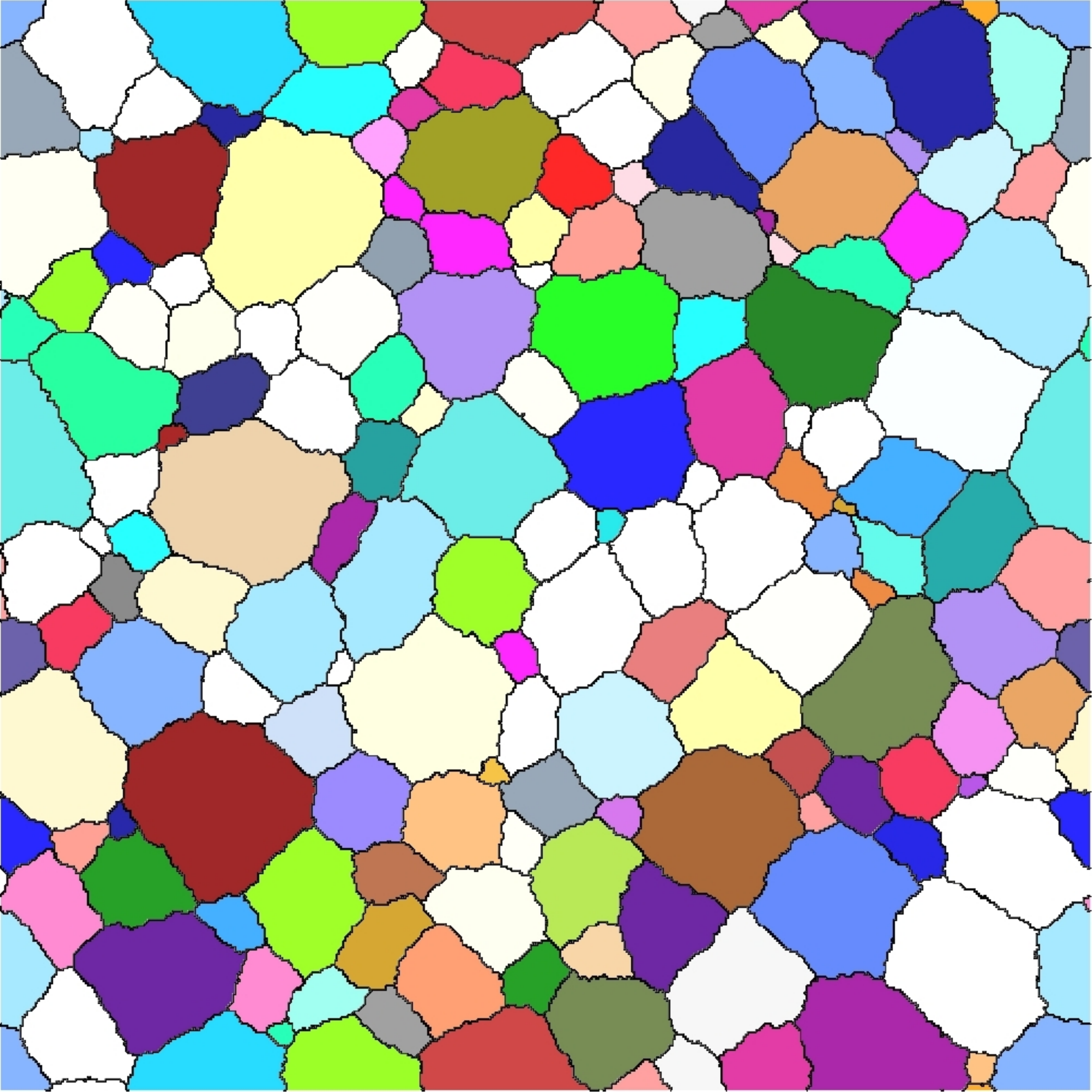}}
\caption{Grain growth simulation via kinetic Monte Carlo (SPPARKS).}
\label{fig:spparks}
\end{figure}

\subsection{Results}

\begin{figure}
\centering
\includegraphics[width=0.5\textwidth]{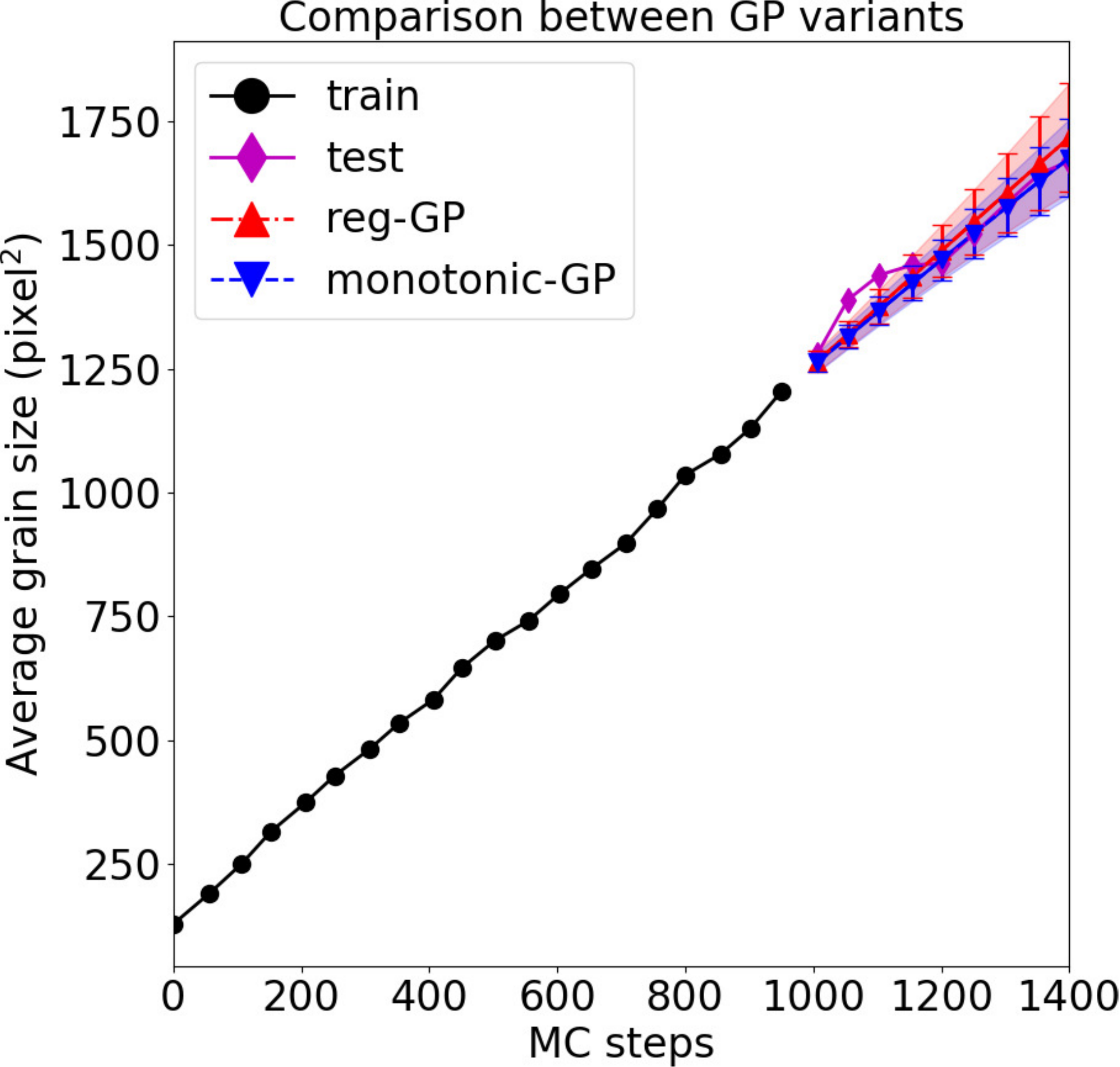}
\caption{Comparison between regular GP and monotonic GP in extrapolating QoI in the SPPARKS grain growth simulation for $k_B T_s=0.65$. Shaded regions correspond to 90\% confidence intervals $\mu \pm 1.645 \sigma$.}
\label{fig:kT_065_extrapolation}
\end{figure}

\begin{figure}
\centering
\includegraphics[width=0.5\textwidth]{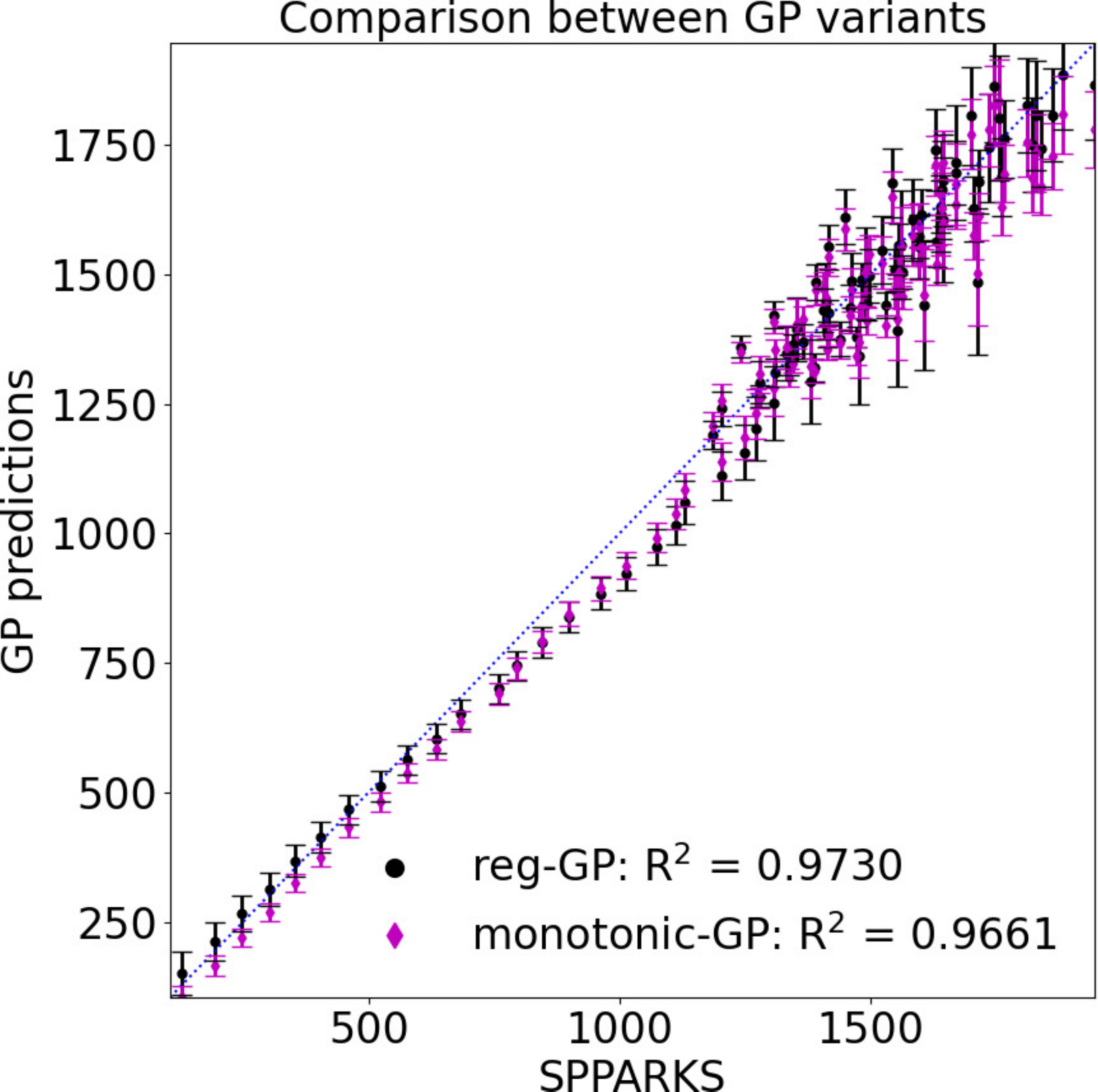}
\caption{Comparison of testing dataset predictions between the regular and monotonic GPs in kinetic Monte Carlo SPPARKS grain growth simulation. The output is the average grain size, whereas the input is Monte Carlo step $t$. An error bar of $\mu \pm 1.645 \sigma$, which corresponds to a 90\% confidence interval is shown. The dashed line $y=x$ is presented as a reference.}
\label{fig:spparks_comparison2}
\end{figure}

Using SPPARKS as the forward model, we build a dataset with $k_B T_s \in \{ 0.15, 0.25, 0.35, 0.45, 0.55, 0.65, 0.75, 0.85\}$ at multiple snapshots in time $t \in \{$ 0.000, 56.879, 106.899, 152.482, 206.331, 251.886, 306.530, 353.403, 406.482, 450.629, 502.591, 554.926, 603.345, 653.470, 706.620, 756.473, 800.112, 855.231, 901.232, 950.060, 1005.970, 1054.070, 1102.890, 1155.320, 1200.360, 1251.300, 1302.020, 1353.410, 1400.000$\}$ Monte Carlo steps. In this case study, the average grain size, measured in pixel$^2$, is used as the quantity of interest. Similar to Tran et al.~\cite{tran2020an}, we apply a filter of 50 pixel$^2$ to eliminate small grains before computing the average grain size. The training and testing datasets are divided as follows: if $k_B T_s \leq 0.75$ or if $t < 1000$, the data point belongs to the training dataset; otherwise, it belongs to the testing dataset. Under this condition, a training dataset of 140 data points and a testing dataset of 92 data points are obtained.
The inputs of the GP are $(k_B T_s, t)$, and the output of the GP is the average grain size.

It is noted that the data is naturally monotonic, as grains grow in time, which is shown in Figure~\ref{fig:spparks}. At the boundary for the training and testing datasets, at $t = 950.060$ Monte Carlo step, for $k_B T_s = 0.65$, the average grain size is 1203.6605 pixel$^2$,
whereas at $t = 1005.970$ Monte Carlo step, for $k_B T_s = 0.65$, the average grain size is 1279.0099 pixel$^2$. This is clearly shown in Figure~\ref{fig:spparks_comparison2}, where the testing predictions are mostly more than 1250 pixel$^2$.

Figure~\ref{fig:kT_065_extrapolation} compares the regular GP and the monotonic GP in terms of extrapolating beyond the training regime for $k_B T_s = 0.65$, showing a consistent behavior between the monotonic GP and the regular GP. The posterior means between the regular GP and the classical GP are almost exactly the same. However, the posterior variance is significantly reduced in the monotonic GP, which could be explained by $M$ extra inducing points for monotonicity classification, as described in Section~\ref{subsec:monotonicGP}.

Figure~\ref{fig:spparks_comparison2} compares the regular GP and the monotonic GP in terms of accuracy with the error bar of $\mu \pm 1.645 \sigma$, showing that the regular GP ($R^2 = 0.9730$) is slightly better than the monotonic GP ($R^2 = 0.9661$). It can be explained that imposing monotonicity to the GP formulation comes at a small cost for accuracy.

\section{Case study: Strain-rate-dependent stress-strain with crystal plasticity finite element}
\label{sec:ex:damask}

\subsection{Model description}

We adopt and extend the dataset from one of our previous studies~\cite{tran2020solving}, generated from Fe-22Mn-0.6C TWIP steel with the dislocation-density-based constitutive model. The material parameters are as described in Steinmetz et al.~\cite{steinmetz2013revealing} and summarized in Section 6.2.3 and Tables 8 and 9 in \cite{roters2019damask}.
The constitutive model was validated experimentally by Wong et al~\cite{wong2016crystal}, and for the sake of completeness, we briefly summarize the constitutive model here.
The TWIP/TRIP steel constitutive model is parameterized in terms of dislocation density, $\varrho$, the dipole dislocation density, $\varrho_{\text{di}}$, the twin volume fraction, $f_{\text{tw}}$, and the $\varepsilon$-martensite volume fraction, $f_{\text{tr}}$.
A model for the plastic velocity gradient with contribution of mechanical twinning and phase transformation was developed in Kalidindi~\cite{kalidindi1998incorporation} and is given by
\begin{equation}
\begin{array}{lll}
\mathbf{L}_\text{p} &=& (1 - f^{\text{tot}}_{\text{tw}} -f^{\text{tot}}_{\text{tr}}) \big( \sum_{\alpha=1}^{N_{\text{s}}} \dot{\gamma}^{\alpha} \mathbf{s}_{\text{s}}^{\alpha} \otimes \mathbf{n}_{\text{s}}^{\alpha} \\
&&+ \sum_{\beta=1}^{N_{\text{tw}}} \dot{\gamma} \mathbf{s}_{\text{tw}} \otimes \mathbf{n}_{\text{tw}}^{\beta} + \sum_{\chi=1}^{N_{\text{tr}}} \dot{\gamma}^{\chi} \mathbf{s}_{\text{tr}} \otimes \mathbf{n}_{\text{tr}}^{\beta} \big),
\end{array}
\end{equation}
where $\chi = 1, \dots,N_{\text{tr}}$ is the $\varepsilon$-martensite with volume fraction $f_{\text{tr}}$ on $N_{\text{tr}}$ transformation systems, $\mathbf{s}_{\text{s}}^{\alpha}$ and $\mathbf{n}_{\text{s}}^{\alpha}$ are unit vectors along the shear direction and shear plane normal of $N_{\text{s}}$ slip systems $\alpha$, $\mathbf{s}_{\text{tw}}^{\alpha}$ and $\mathbf{n}_{\text{tw}}^{\alpha}$ are those of $N_{\text{tw}}$ twinning systems $\beta$, and $\mathbf{s}_{\text{tw}}^{\alpha}$ and $\mathbf{n}_{\text{tr}}^{\alpha}$ are those of $N_{\text{tr}}$ transformation systems $\chi$.
The Orowan equation models the shear rate on the slip system $\alpha$ as
\begin{equation}
\dot{\gamma}^{\alpha} = \rho_e b_{\text{s}} \nu_0 \exp \left[ - \frac{Q}{k_B T} \left\{ 1 - \left( \frac{\tau_{\text{eff}}^{\alpha}}{\tau_{\text{sol}}} \right)^p \right\}^q \right],
\end{equation}
where $b_{\text{s}}$ is the length of the slip Burgers vector, $\nu_0$ is a reference velocity, $Q_{\text{s}}$ is the activation energy for slip, $k_B$ is the Boltzmann constant, $T$ is the temperature, $\tau_{\text{eff}}$ is the effective resolved shear stress, $\tau_{\text{sol}}$ is the solid solution strength, and $0 < \rho_\text{s} \leq 1$ and $1 \leq q_\text{s} \leq 2 $ are fitting parameters controlling the glide resistance profile.
Blum and Eisenlohr~\cite{blum2009dislocation} model the evolution of dislocation densities, particularly the generation of unipolar dislocation density and formation of dislocation dipoles, respectively, as
\begin{equation}
\dot{\varrho} = \frac{|\dot{\gamma}|}{b_\text{s}} \Gamma_\text{s} - \frac{2\hat{d}{b_\text{s}}} \varrho |\dot{\gamma}|,
\end{equation}
\begin{equation}
\dot{\varrho}_\text{di} = \frac{2(\hat{d} - \widecheck{d})}{b_\text{s}} \varrho |\dot{\gamma}| - \frac{2 \widecheck{d}}{b_\text{s}} \varrho_\text{di} |\dot{\gamma}| - \varrho_\text{di} \frac{4\nu_\text{cl}}{\hat{d} - \widecheck{d}},
\end{equation}
where the glide distance below which two dislocations form a stable dipole is
\begin{equation}
\hat{d} = \frac{3G b_\text{s}}{16\pi |\tau|},
\end{equation}
$\widecheck{d} = D_a b_\text{s}$ is the distance below which two dislocations annihilate, and
the dislocation climb velocity is
\begin{equation}
\nu_\text{cl} = \frac{G D_0 V_\text{cl}}{\pi (1 - \nu) k_\text{B} T} \frac{1}{\hat{d} + \widecheck{d}}\exp \left( - \frac{Q_\text{cl}}{k_\text{B} T} \right).
\end{equation}
Here,
$D_0$ is the pre-factor of self-diffusion coefficient,
$V_\text{cl}$ is the activation volume for climb, and
$Q_\text{cl}$ is the activation energy for climb.
Strain hardening is described in terms of a dislocation mean free path, where the mean free path is denoted by $\Gamma$.
The mean free path for slip is modeled as
\begin{equation}
\frac{1}{\Gamma_\text{s}} = \frac{1}{D} + \frac{1}{\lambda_\text{s}} + \frac{1}{\lambda_\text{tw}} + \frac{1}{\lambda_\text{tr}}
\end{equation}
where $D$ is the average grain size and
\begin{equation}
\frac{1}{\lambda_\text{s}^{\alpha}} = \frac{1}{i_\text{s}} \left( \sum_{\alpha'=1}^{N_\text{s}} p^{\alpha \alpha'} (\varrho^{\alpha'} + \varrho_\text{di}^{\alpha'}) \right)^{1/2},
\end{equation}
\begin{equation}
\frac{1}{\lambda_{\text{tw}}^{\alpha}} = \sum_{\beta = 1}^{N_{\text{tw}}} h^{\alpha \beta} \frac{f^\beta_{\text{tw}}}{t_\text{tw} (1 - f^\text{tot}_\text{tw})}, \quad
\frac{1}{\lambda_\text{tr}^{\alpha}} = \sum_{\chi=1}^{N_\text{tr}} h^{\alpha \chi} \frac{f^{\chi}_\text{tr}}{t_\text{tr}(1 - f^\text{tot}_\text{tr})}.
\end{equation}
Here, $i_\text{s}$ is a fitting parameter, $t_\text{tw}$ is the average twin thickness, and $t_\text{tr}$ is the average $\varepsilon$-martensite thickness.
The mean free path for twinning and for transformation are computed, respectively, as
\begin{equation}
\frac{1}{\Gamma^\beta_\text{tw}} = \frac{1}{i_\text{tw}} \left( \frac{1}{D} + \sum_{\beta'=1}^{N_\text{tw}} h^{\beta \beta'} f_{\text{tw}}^{\beta'} \frac{1}{t_\text{tw} ( 1 - f^\text{tot}_\text{tw})} \right),
\end{equation}
\begin{equation}
\frac{1}{\Gamma^\chi_\text{tr}} = \frac{1}{i_\text{tr}} \left( \frac{1}{D} + \sum_{\chi'=1}^{N_\text{tr}} h^{\chi \chi'} f_{\text{tr}}^{\chi'} \frac{1}{t_\text{tr} ( 1 - f^\text{tot}_\text{tr})} \right),
\end{equation}
with $i_\text{w}$ and $i_\text{tr}$ being fitting parameters.
The nucleation rates for twins and $\varepsilon$-martensite are $\dot{N} = \dot{N}_0 P_\text{ncs} P$, where the probability $P$ that a nucleus bows out to form a twin or $\varepsilon$-martensite is
\begin{equation}
p_\text{tw} = \exp \left[ - \left( \frac{\hat{\tau}_\text{tw}}{\tau} \right)^{p_\text{tw}} \right], \quad
p_\text{tr} = \exp \left[ - \left( \frac{\hat{\tau}_\text{tr}}{\tau} \right)^{p_\text{tr}} \right],
\end{equation}
$p_\text{tw}$ and $p_\text{tr}$ are fitting parameters. For more details, readers are referred to Roters et al. \cite{roters2019damask} (cf. Section 6.2.3) and Wong et al \cite{wong2016crystal}.

\begin{figure}
\centering
\subcaptionbox{The microstructure statistical volume element with equiaxed grains investigated in this case study.
\label{fig:samplesve}
}
  [0.45\linewidth]{\includegraphics[width=0.24\textwidth,keepaspectratio]{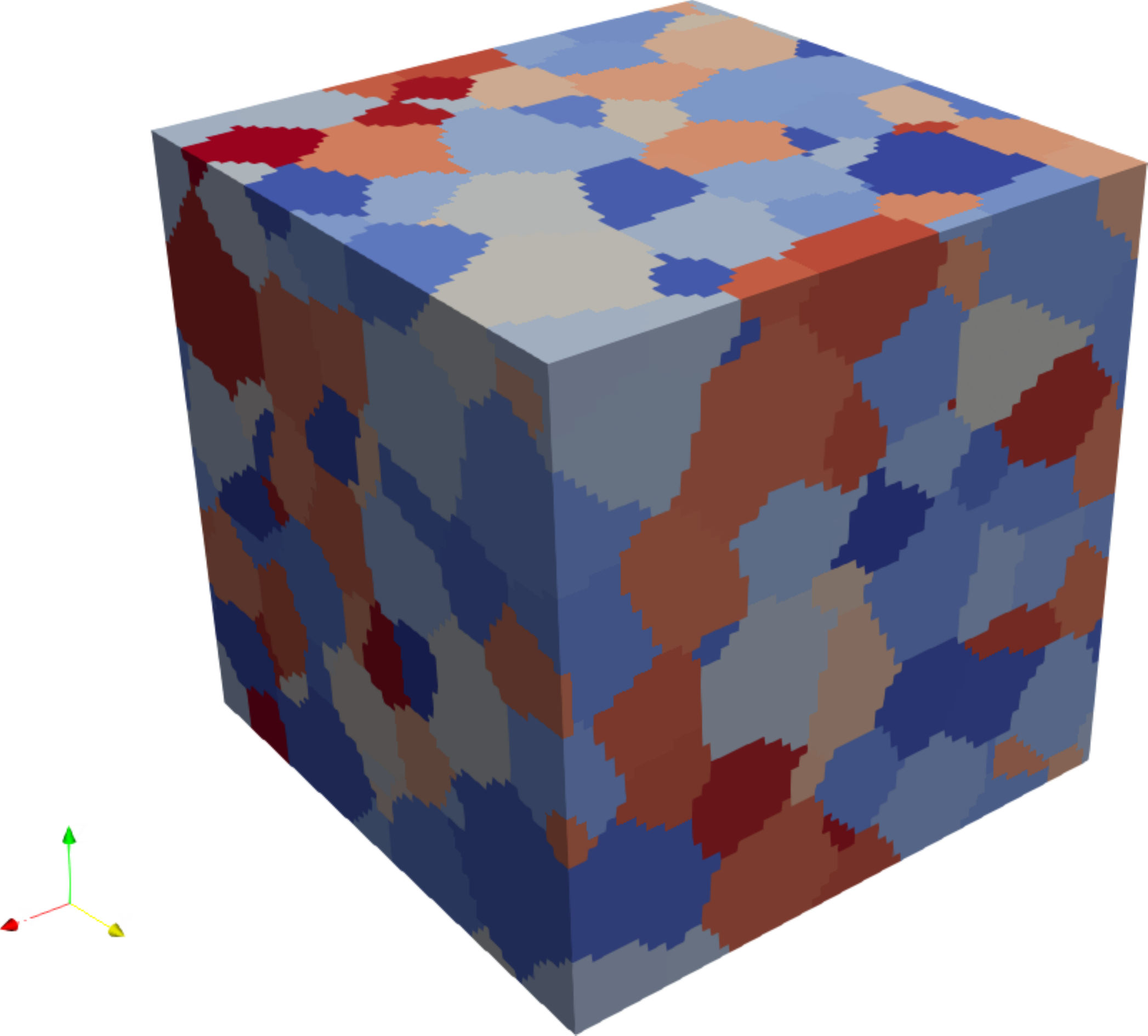}}
\hfill
\subcaptionbox{A snapshot of von Mises stress for $D=10^{-6}m$, $\dot{\varepsilon} = 10^{-4}$ s$^{-1}$ before homogenization.
\label{fig:stresssve}
}
  [0.45\linewidth]{\includegraphics[width=0.24\textwidth, keepaspectratio]{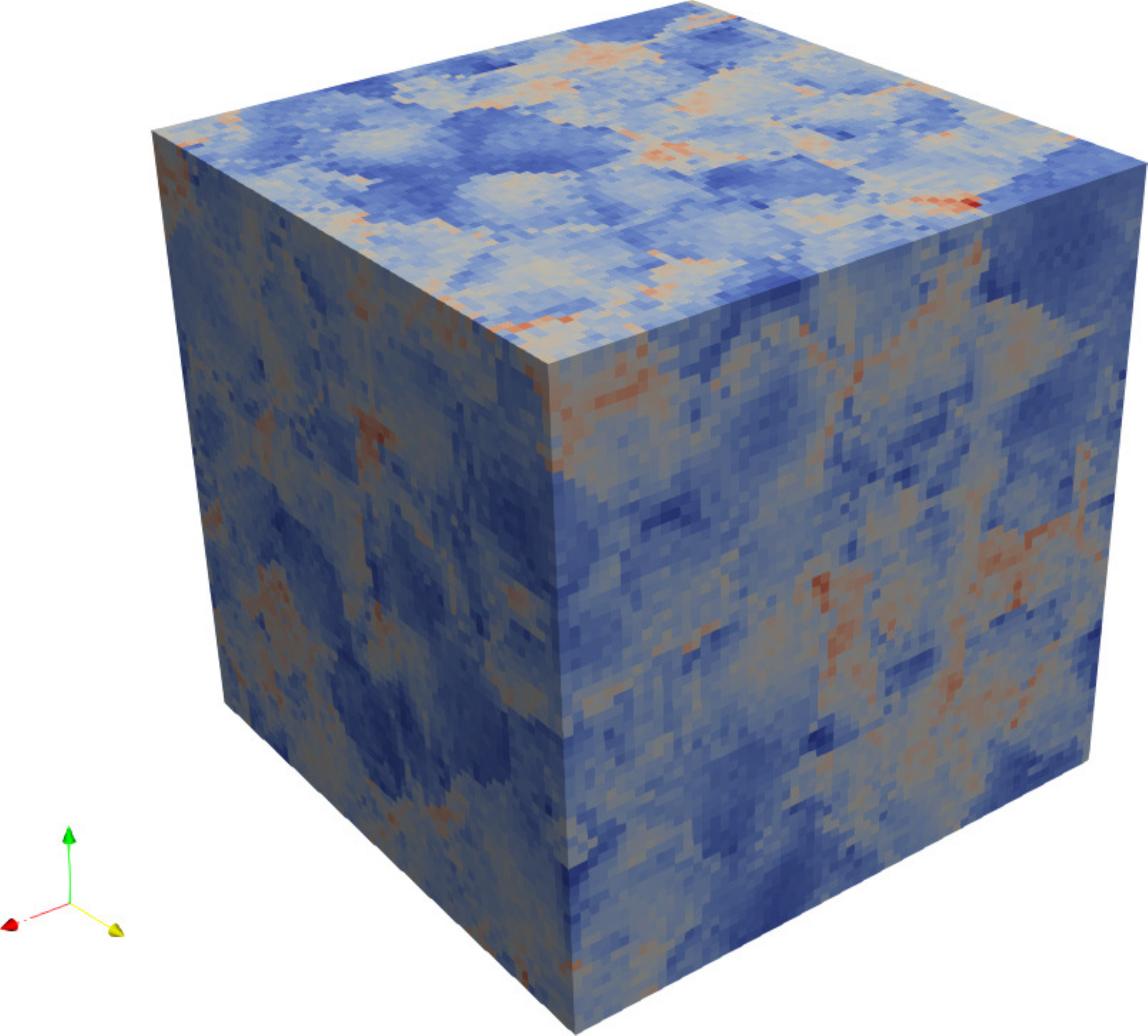}}
\caption{A crystal plasticity finite element example by DAMASK.}
\label{fig:damask}
\end{figure}

Figure~\ref{fig:damask} shows the equiaxed microstructure used in this case study, which has been generated using DREAM.3D~\cite{groeber2014dream} under random crystallographic textures. DAMASK~\cite{diehl2017identifying,roters2019damask} is used as a crystal plasticity finite element, where PETSc~\cite{abhyankar2018petsc,balay2019petsc} is used as the underlying spectral solver.



\subsection{Results}

\begin{figure*}
\centering
\includegraphics[width=1.0\textwidth,keepaspectratio]{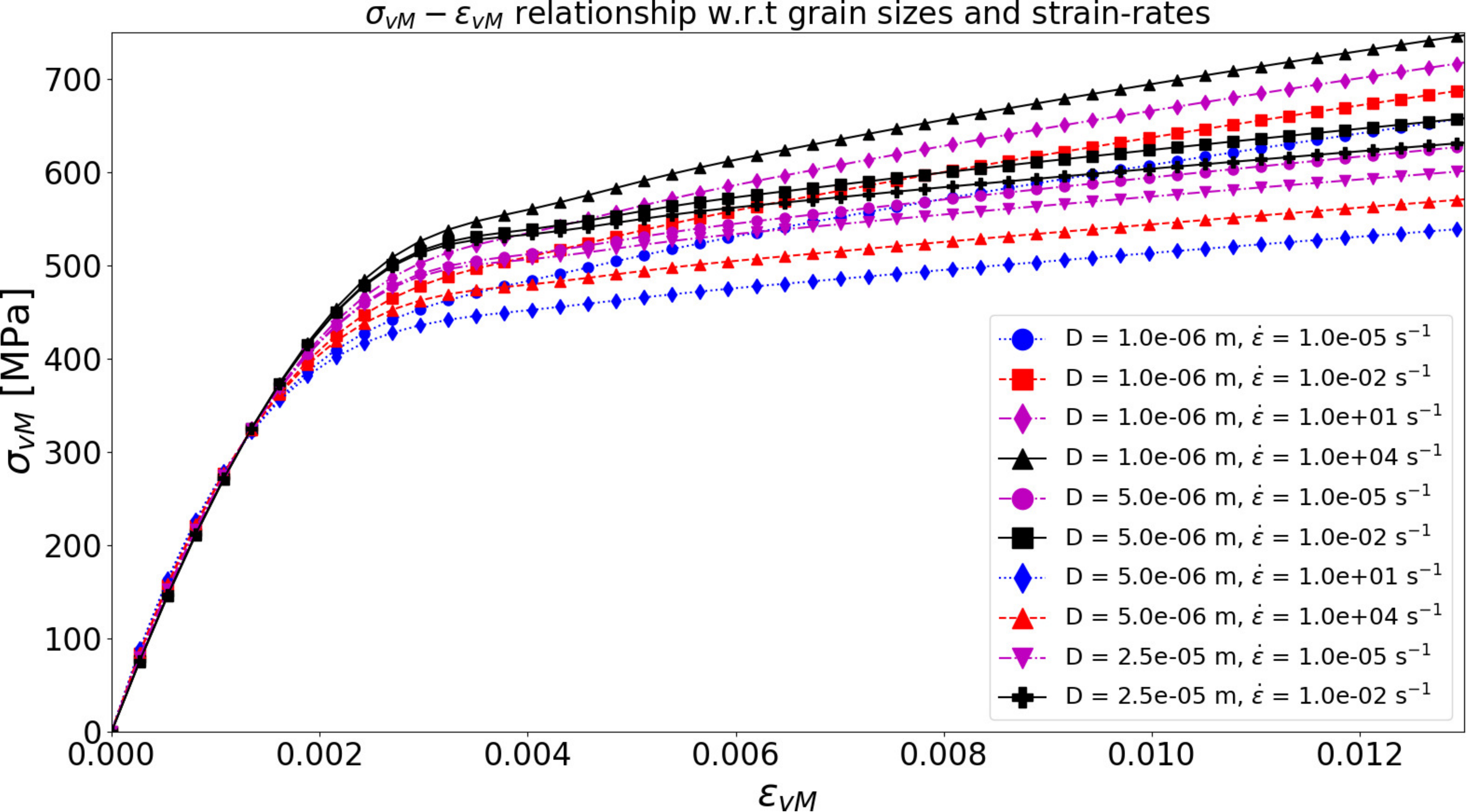}
\caption{Homogenized stress-strain response with different $(D,\dot{\varepsilon},\varepsilon)$ for a fixed microstructure shown in Figure~\ref{fig:damask}. $\sigma$ is monotonically increasing when either $D$ decreases, or $\dot{\varepsilon}$ increases, or $\varepsilon$ increases (within the specified $\varepsilon$ domain).}
\label{fig:stressStrainCollection}
\end{figure*}

\begin{figure}
\centering
\includegraphics[width=0.5\textwidth,keepaspectratio]{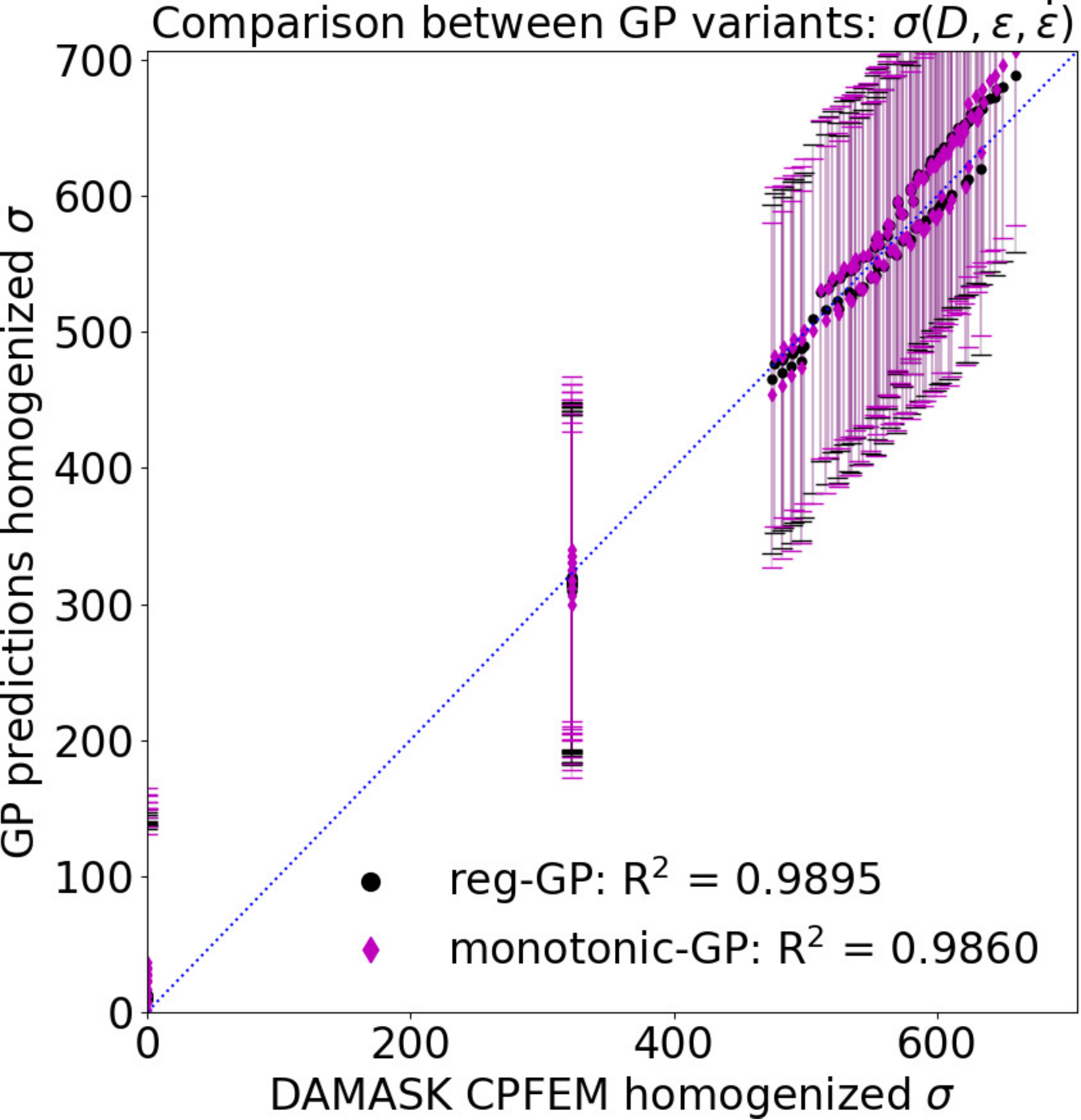}
\caption{Comparison of testing dataset predictions between the regular and monotonic GPs in crystal plasticity finite element DAMASK. The output is the homogenized stress $\sigma$, whereas the inputs are average grain size, strain-rate, and strain, respectively, $(D,\dot{\varepsilon},\varepsilon)$. }
\label{fig:gp_damask_comparison}
\end{figure}

In this study, we focus on the effect of strain-rate on the stress-strain curve with various average grain sizes. The effect of strain-rate has been experimentally validated in Benzing et al~\cite{benzing2018effects}, and some numerical studies using crystal plasticity finite element model have been done as well, such as those found in Singh et al~\cite{singh2021investigation}.
The inputs of the GP are the average grain size, the strain-rate, the strain, i.e. $(D, \dot{\varepsilon}, \varepsilon)$, and the output of the GP is the stress $\sigma$.
We vary independently the average grain size $D \in \{1.0 \cdot 10^{-6}, 2.5 \cdot 10^{-6}, 5.0 \cdot 10^{-6}, 1.0 \cdot 10^{-5}, 2.5 \cdot 10^{-5} \}$ m, as well as the strain-rate $\dot{\varepsilon} \in \{ 10^{-5}, 10^{-4}, 10^{-3}, 10^{-2}, 10^{-1}, 10^{0}, 10^{1}, 10^{2}, 10^{3}, 10^{4} \}$ s$^{-1}$. Numerically unstable results are removed in the post-process.
Some examples of homogenized stress-strain responses are shown in Figure~\ref{fig:stressStrainCollection} for visualization purpose.

The training and testing datasets are divided as follows: if the average grain size $D$ is less than $5\cdot 10^{-6}$m or if the strain-rate $\dot{\varepsilon} = 10^1$ s$^{-1}$, the data point belongs to the training dataset; otherwise, it belongs to the testing dataset. Under this splitting condition, a training dataset of 340 data points and a testing dataset of 120 data points are obtained.

Figure~\ref{fig:gp_damask_comparison} shows the accuracy comparison between the regular GP and the classical GP. The $R^2$ coefficient for the regular GP is 0.9895, whereas the $R^2$ coefficient for the monotonic GP is slightly lower, 0.9860. Both GPs do very well, but the \textcolor{black}{monotonic GP is slightly worse}. The regular GP predictions are also more consistent in the large stress regime $\sigma$.

\section{Discussion and Conclusion}
\label{sec:discussion}

In this paper, we compare the monotonic GP against the regular GP for various materials datasets, namely fatigue, grain growth, and homogenized stress-strain behaviors. The fatigue dataset is the only experimental dataset with noise, whereas the other two datasets, i.e. grain growth and homogenized stress-strain, are computationally generated by SPPARKS and DAMASK, respectively.
We test out their predictive capabilities in both interpolation and extrapolation, but are mainly concerned with extrapolation.

For the fatigue example featured in Section~\ref{sec:ex:fatigue}, the experimental data does not adhere strictly to theoretical derivations.
This is where the monotonic GP outperforms the regular GP significantly.
In the other two examples, which featured datasets generated by computational models, both GPs perform very closely with each other, even though the regular GP performs slightly better than the monotonic GP.
This can be explained by imposing the monotonicity on the regular GP comes a small cost for accuracy and numerical performance.
It is also noted that the monotonic behavior in the monotonic GP is only guaranteed within the training domain. Beyond the training regime, the monotonicity effect imposed by introducing $M$ inducing points fades away: the further the monotonic GP goes beyond the training regime, the more likely it is the GP will violate the monotonicity constraints.
Last but not least, the posterior variances of the monotonic GP are significantly smaller than the posterior variances of the regular GP, while the posterior means are roughly the same.

Constructing an accurate and physically faithful surrogate model is important for uncertainty quantification~\cite{acar2021recent}. Computationally efficient surrogate models are commonly used, for example, in sampling~\cite{tallman2019gaussian,tallman2020uncertainty} or to solve a stochastic inverse problem in structure-property relationship~\cite{tran2020solving,tran2021solving1}. Monotonicity is common in materials science, such as the famous Hall-Petch relationship~\cite{cordero2016six}, i.e. $\sigma_y = \sigma_0 + k \frac{1}{\sqrt{D}}$. Exploiting such physical constraints and insights may improve the efficiency of machine learning in the future.

\textcolor{black}{The objective of this work is to examine the performance of the monotonic GP compared to the regular GP in different settings, with an emphasis on materials science. Our conclusion is that the monotonic GP is best suited for sparse and noisy settings, which is fairly common in experimental materials science, where the data acquisition process is expensive (and intensive in terms of both time and labor perspectives) and the data is often corrupted by the noise induced by the finite-size effect of microstructures.}

\textcolor{black}{
It is worthy to note that the monotonic GP indeed performs slightly worse in the second and third case studies.
In these case studies, both the training and testing datasets are already monotonic.
With the use of linear basis function for the global trend, the regular GP correctly picks up the monotonic behavior, and therefore, is able to make accurate prediction.
While the monotonic GP shares the same settings with the regular GP, there are some approximations involved in constructing the joint covariance matrix.
In ideally monotonic cases such as the second and third examples, the covariance matrix should asymptotically reduce to the diagonal block matrix as $\nu \to 0$
\begin{equation}
\mathbf{K}_{\text{joint}}
= \begin{bmatrix} \mathbf{K}_{\mathbf{f}, \mathbf{f}} & \mathbf{K}_{\mathbf{f}, \mathbf{f}'} \\ \mathbf{K}_{\mathbf{f}', \mathbf{f}} & \mathbf{K}_{\mathbf{f}', \mathbf{f}'} \end{bmatrix}
\xrightarrow[\nu \to 0]{}  \begin{bmatrix} \mathbf{I}_{\mathbf{f}, \mathbf{f}} & \mathbf{0} \\ \mathbf{0} & \mathbf{I}_{\mathbf{f}', \mathbf{f}'} \end{bmatrix},
\end{equation}
which points to how the step function is approximated. In practice, a small numerical degradation is expected as $\nu$ cannot be smaller than machine epsilon. In the monotonic GP approach adopted in this paper, the step function is approximated by a probit function described in Equation~\ref{eq:probitMonotonicity}. It might be possible to increase the performance of the monotonic GP by even further reducing the $\nu$ parameter. This may lead to an improvement in performance for the monotonic GP with monotonic datasets, but might also lead to other numerical conditioning problems with respect to the inverse of the joint covariance matrix $\mathbf{K}_\text{joint}$.
We conclude that for datasets that are already monotonic, applying a monotonic GP formulation may be unnecessary and may lead to a slight degradation in performance.
For datasets that are sparse and noisy, imposing monotonicity when it is appropriate may result in a substantial improvement in performance.
}

\section*{Acknowledgment}
%

This work was supported by the Laboratory Directed Research and Development program at Sandia National Laboratories, a multimission laboratory managed and operated by National Technology and Engineering Solutions of Sandia LLC, a wholly owned subsidiary of Honeywell International Inc. for the U.S. Department of Energy’s National Nuclear Security Administration under contract DE-NA0003525.

%
%

\bibliographystyle{asmems4}
\bibliography{lib}   

\begin{thebibliography}{10}

\bibitem{national2011materials}
{National Science and Technology Council (US)}, 2011.
\newblock {\em {Materials Genome Initiative} for global competitiveness}.
\newblock Executive Office of the President, National Science and Technology
  Council.

\bibitem{cordero2016six}
Cordero, Z.~C., Knight, B.~E., and Schuh, C.~A., 2016.
\newblock ``Six decades of the {Hall--Petch} effect--a survey of grain-size
  strengthening studies on pure metals''.
\newblock {\em International Materials Reviews, {\bf 61}}(8), pp.~495--512.

\bibitem{tallman2019gaussian}
Tallman, A.~E., Stopka, K.~S., Swiler, L.~P., Wang, Y., Kalidindi, S.~R., and
  McDowell, D.~L., 2019.
\newblock ``Gaussian-process-driven adaptive sampling for reduced-order
  modeling of texture effects in polycrystalline alpha-{Ti}''.
\newblock {\em JOM, {\bf 71}}(8), pp.~2646--2656.

\bibitem{tallman2020uncertainty}
Tallman, A.~E., Swiler, L.~P., Wang, Y., and McDowell, D.~L., 2020.
\newblock ``Uncertainty propagation in reduced order models based on crystal
  plasticity''.
\newblock {\em Computer Methods in Applied Mechanics and Engineering, {\bf
  365}}, p.~113009.

\bibitem{yabansu2019application}
Yabansu, Y.~C., Iskakov, A., Kapustina, A., Rajagopalan, S., and Kalidindi,
  S.~R., 2019.
\newblock ``Application of {Gaussian process} regression models for capturing
  the evolution of microstructure statistics in aging of nickel-based
  superalloys''.
\newblock {\em Acta Materialia, {\bf 178}}, pp.~45--58.

\bibitem{tran2020solving}
Tran, A., and Wildey, T., 2020.
\newblock ``Solving stochastic inverse problems for property-structure linkages
  using data-consistent inversion and machine learning''.
\newblock {\em JOM, {\bf 73}}, pp.~72--89.

\bibitem{tran2021solving1}
Tran, A., and Wildey, T., 2021.
\newblock ``Solving stochastic inverse problems for property-structure linkages
  using data-consistent inversion''.
\newblock In TMS 2021 150th Annual Meeting \& Exhibition, Springer, pp.~1--8.

\bibitem{tran2020multi}
Tran, A., Tranchida, J., Wildey, T., and Thompson, A.~P., 2020.
\newblock ``Multi-fidelity machine-learning with uncertainty quantification and
  {B}ayesian optimization for materials design: {A}pplication to ternary random
  alloys''.
\newblock {\em The Journal of Chemical Physics, {\bf 153}}, p.~074705.

\bibitem{khatamsaz2021efficiently}
Khatamsaz, D., Molkeri, A., Couperthwaite, R., James, J., Arr{\'o}yave, R.,
  Allaire, D., and Srivastava, A., 2021.
\newblock ``Efficiently exploiting process-structure-property relationships in
  material design by multi-information source fusion''.
\newblock {\em Acta Materialia, {\bf 206}}, p.~116619.

\bibitem{fernandez2018use}
Fern{\'a}ndez-Godino, M.~G., Balachandar, S., and Haftka, R., 2018.
\newblock ``On the use of symmetries in building surrogate models''.
\newblock {\em Journal of Mechanical Design}.

\bibitem{jidling2017linearly}
Jidling, C., Wahlstr{\"o}m, N., Wills, A., and Sch{\"o}n, T.~B., 2017.
\newblock ``Linearly constrained {G}aussian processes''.
\newblock {\em arXiv preprint arXiv:1703.00787}.

\bibitem{agrell2019gaussian}
Agrell, C., 2019.
\newblock ``Gaussian processes with linear operator inequality constraints''.
\newblock {\em arXiv preprint arXiv:1901.03134}.

\bibitem{lange2021linearly}
Lange-Hegermann, M., 2021.
\newblock ``Linearly constrained {G}aussian processes with boundary
  conditions''.
\newblock In International Conference on Artificial Intelligence and
  Statistics, PMLR, pp.~1090--1098.

\bibitem{swiler2020survey}
Swiler, L.~P., Gulian, M., Frankel, A.~L., Safta, C., and Jakeman, J.~D., 2020.
\newblock ``A survey of constrained gaussian process regression: {A}pproaches
  and implementation challenges''.
\newblock {\em Journal of Machine Learning for Modeling and Computing, {\bf
  1}}(2).

\bibitem{riihimaki2010gaussian}
Riihim{\"a}ki, J., and Vehtari, A., 2010.
\newblock ``Gaussian processes with monotonicity information''.
\newblock In Proceedings of the thirteenth international conference on
  artificial intelligence and statistics, JMLR Workshop and Conference
  Proceedings, pp.~645--652.

\bibitem{golchi2015monotone}
Golchi, S., Bingham, D.~R., Chipman, H., and Campbell, D.~A., 2015.
\newblock ``Monotone emulation of computer experiments''.
\newblock {\em SIAM/ASA Journal on Uncertainty Quantification, {\bf 3}}(1),
  pp.~370--392.

\bibitem{ustyuzhaninov2020monotonic}
Ustyuzhaninov, I., Kazlauskaite, I., Ek, C.~H., and Campbell, N., 2020.
\newblock ``Monotonic {G}aussian process flows''.
\newblock In International Conference on Artificial Intelligence and
  Statistics, PMLR, pp.~3057--3067.

\bibitem{pensoneault2020nonnegativity}
Pensoneault, A., Yang, X., and Zhu, X., 2020.
\newblock ``Nonnegativity-enforced {G}aussian process regression''.
\newblock {\em Theoretical and Applied Mechanics Letters, {\bf 10}}(3),
  pp.~182--187.

\bibitem{tan2017monotonic}
Tan, M. H.~Y., 2017.
\newblock ``Monotonic metamodels for deterministic computer experiments''.
\newblock {\em Technometrics, {\bf 59}}(1), pp.~1--10.

\bibitem{chen2021solving}
Chen, Y., Hosseini, B., Owhadi, H., and Stuart, A.~M., 2021.
\newblock ``Solving and learning nonlinear {PDE}s with {G}aussian processes''.
\newblock {\em Journal of Computational Physics, {\bf 447}}, p.~110668.

\bibitem{rasmussen2006gaussian}
Rasmussen, C.~E., 2006.
\newblock {\em Gaussian processes in machine learning}.
\newblock {MIT Press}.

\bibitem{shahriari2016taking}
Shahriari, B., Swersky, K., Wang, Z., Adams, R.~P., and de~Freitas, N., 2016.
\newblock ``Taking the human out of the loop: A review of {B}ayesian
  optimization''.
\newblock {\em Proceedings of the IEEE, {\bf 104}}(1), pp.~148--175.

\bibitem{vanhatalo2012bayesian}
Vanhatalo, J., Riihim{\"a}ki, J., Hartikainen, J., Jyl{\"a}nki, P., Tolvanen,
  V., and Vehtari, A., 2012.
\newblock ``{Bayesian modeling with Gaussian processes using the GPstuff
  toolbox}''.
\newblock {\em arXiv preprint arXiv:1206.5754}.

\bibitem{vanhatalo2013gpstuff}
Vanhatalo, J., Riihim{\"a}ki, J., Hartikainen, J., Jyl{\"a}nki, P., Tolvanen,
  V., and Vehtari, A., 2013.
\newblock ``{GPstuff: Bayesian modeling with Gaussian processes}''.
\newblock {\em Journal of Machine Learning Research, {\bf 14}}(Apr),
  pp.~1175--1179.

\bibitem{tran2020smfbo2cogp}
Tran, A., Wildey, T., and McCann, S., 2020.
\newblock ``{sMF-BO-2CoGP: A sequential multi-fidelity constrained Bayesian
  optimization for design applications}''.
\newblock {\em Journal of Computing and Information Science in Engineering,
  {\bf 20}}(3), pp.~1--15.

\bibitem{yang2020bifidelity}
Yang, X., Zhu, X., and Li, J., 2020.
\newblock ``When bifidelity meets {CoKriging}: An efficient physics-informed
  multifidelity method''.
\newblock {\em SIAM Journal on Scientific Computing, {\bf 42}}(1),
  pp.~A220--A249.

\bibitem{xiao2018extended}
Xiao, M., Zhang, G., Breitkopf, P., Villon, P., and Zhang, W., 2018.
\newblock ``Extended co-kriging interpolation method based on multi-fidelity
  data''.
\newblock {\em Applied Mathematics and Computation, {\bf 323}}, pp.~120--131.

\bibitem{minka2001expectation}
Minka, T.~P., 2001.
\newblock ``Expectation propagation for approximate {B}ayesian inference''.
\newblock In Proceedings of the Seventeenth Conference on Uncertainty in
  Artificial Intelligence, UAI’01, AUAI Press, pp.~362--369.

\bibitem{kuss2005assessing}
Kuss, M., Rasmussen, C.~E., and Herbrich, R., 2005.
\newblock ``Assessing approximate inference for binary gaussian process
  classification.''.
\newblock {\em Journal of Machine Learning Research, {\bf 6}}(10).

\bibitem{counts2008predicting}
Counts, W.~A., Braginsky, M.~V., Battaile, C.~C., and Holm, E.~A., 2008.
\newblock ``Predicting the {Hall--Petch} effect in fcc metals using non-local
  crystal plasticity''.
\newblock {\em International journal of plasticity, {\bf 24}}(7),
  pp.~1243--1263.

\bibitem{fernandes2000further}
Fernandes, J., and Vieira, M., 2000.
\newblock ``Further development of the hybrid model for polycrystal
  deformation''.
\newblock {\em Acta materialia, {\bf 48}}(8), pp.~1919--1930.

\bibitem{karolczuk2022application}
Karolczuk, A., and S{\l}o{\'n}ski, M., 2022.
\newblock ``Application of the {G}aussian process for fatigue life prediction
  under multiaxial loading''.
\newblock {\em Mechanical Systems and Signal Processing, {\bf 167}}, p.~108599.

\bibitem{karolczuk2020application}
Karolczuk, A., and Kluger, K., 2020.
\newblock ``Application of life-dependent material parameters to lifetime
  calculation under multiaxial constant-and variable-amplitude loading''.
\newblock {\em International Journal of Fatigue, {\bf 136}}, p.~105625.

\bibitem{arroyave2019systems}
Arr{\'o}yave, R., and McDowell, D.~L., 2019.
\newblock ``Systems approaches to materials design: {Past}, present, and
  future''.
\newblock {\em Annual Review of Materials Research, {\bf 49}}(1), pp.~103--126.

\bibitem{zhu1997algorithm}
Zhu, C., Byrd, R.~H., Lu, P., and Nocedal, J., 1997.
\newblock ``Algorithm 778: {L-BFGS-B: Fortran} subroutines for large-scale
  bound-constrained optimization''.
\newblock {\em ACM Transactions on mathematical software (TOMS), {\bf 23}}(4),
  pp.~550--560.

\bibitem{garcia2008three}
Garcia, A.~L., Tikare, V., and Holm, E.~A., 2008.
\newblock ``Three-dimensional simulation of grain growth in a thermal gradient
  with non-uniform grain boundary mobility''.
\newblock {\em Scripta Materialia, {\bf 59}}(6), pp.~661--664.

\bibitem{plimpton2009crossing}
Plimpton, S., Battaile, C., Chandross, M., Holm, L., Thompson, A., Tikare, V.,
  Wagner, G., Webb, E., Zhou, X., Cardona, C.~G., et~al., 2009.
\newblock ``Crossing the mesoscale no-man’s land via parallel kinetic {Monte
  Carlo}''.
\newblock {\em Sandia Report SAND2009-6226}.

\bibitem{anderson1989computer}
Anderson, M., Grest, G., and Srolovitz, D., 1989.
\newblock ``Computer simulation of normal grain growth in three dimensions''.
\newblock {\em Philosophical Magazine B, {\bf 59}}(3), pp.~293--329.

\bibitem{tran2020an}
Tran, A., Mitchell, J.~A., Swiler, L.~P., and Wildey, T., 2020.
\newblock ``An active-learning high-throughput microstructure calibration
  framework for process-structure linkage in materials informatics''.
\newblock {\em Acta Materialia, {\bf 194}}, pp.~80--92.

\bibitem{steinmetz2013revealing}
Steinmetz, D.~R., J{\"a}pel, T., Wietbrock, B., Eisenlohr, P.,
  Gutierrez-Urrutia, I., Saeed-Akbari, A., Hickel, T., Roters, F., and Raabe,
  D., 2013.
\newblock ``Revealing the strain-hardening behavior of twinning-induced
  plasticity steels: {Theory, simulations, experiments}''.
\newblock {\em Acta Materialia, {\bf 61}}(2), pp.~494--510.

\bibitem{roters2019damask}
Roters, F., Diehl, M., Shanthraj, P., Eisenlohr, P., Reuber, C., Wong, S.~L.,
  Maiti, T., Ebrahimi, A., Hochrainer, T., Fabritius, H.-O., et~al., 2019.
\newblock ``{DAMASK--The D{\"u}sseldorf Advanced Material Simulation Kit for
  modeling multi-physics crystal plasticity, thermal, and damage phenomena from
  the single crystal up to the component scale}''.
\newblock {\em Computational Materials Science, {\bf 158}}, pp.~420--478.

\bibitem{wong2016crystal}
Wong, S.~L., Madivala, M., Prahl, U., Roters, F., and Raabe, D., 2016.
\newblock ``A crystal plasticity model for twinning-and transformation-induced
  plasticity''.
\newblock {\em Acta Materialia, {\bf 118}}, pp.~140--151.

\bibitem{kalidindi1998incorporation}
Kalidindi, S.~R., 1998.
\newblock ``Incorporation of deformation twinning in crystal plasticity
  models''.
\newblock {\em Journal of the Mechanics and Physics of Solids, {\bf 46}}(2),
  pp.~267--290.

\bibitem{blum2009dislocation}
Blum, W., and Eisenlohr, P., 2009.
\newblock ``Dislocation mechanics of creep''.
\newblock {\em Materials Science and Engineering: A, {\bf 510}}, pp.~7--13.

\bibitem{groeber2014dream}
Groeber, M.~A., and Jackson, M.~A., 2014.
\newblock ``{DREAM. 3D: a digital representation environment for the analysis
  of microstructure in 3D}''.
\newblock {\em Integrating materials and manufacturing innovation, {\bf 3}}(1),
  p.~5.

\bibitem{diehl2017identifying}
Diehl, M., Groeber, M., Haase, C., Molodov, D.~A., Roters, F., and Raabe, D.,
  2017.
\newblock ``{Identifying structure--property relationships through DREAM.3D
  representative volume elements and DAMASK crystal plasticity simulations: An
  integrated computational materials engineering approach}''.
\newblock {\em JOM, {\bf 69}}(5), pp.~848--855.

\bibitem{abhyankar2018petsc}
Abhyankar, S., Brown, J., Constantinescu, E.~M., Ghosh, D., Smith, B.~F., and
  Zhang, H., 2018.
\newblock ``{PETSc/TS: A modern scalable ODE/DAE solver library}''.
\newblock {\em arXiv preprint arXiv:1806.01437}.

\bibitem{balay2019petsc}
Balay, S., Abhyankar, S., Adams, M., Brown, J., Brune, P., Buschelman, K.,
  Dalcin, L., Dener, A., Eijkhout, V., Gropp, W., et~al., 2019.
\newblock ``{PETSc} users manual''.

\bibitem{benzing2018effects}
Benzing, J.~T., Poling, W.~A., Pierce, D.~T., Bentley, J., Findley, K.~O.,
  Raabe, D., and Wittig, J.~E., 2018.
\newblock ``Effects of strain rate on mechanical properties and deformation
  behavior of an austenitic {Fe-25Mn-3Al-3Si TWIP-TRIP} steel''.
\newblock {\em Materials Science and Engineering: A, {\bf 711}}, pp.~78--92.

\bibitem{singh2021investigation}
Singh, L., Vohra, S., and Sharma, M., 2021.
\newblock ``Investigation of strain rate behavior of aluminium and {AA2024}
  using crystal plasticity''.
\newblock {\em Materials Today: Proceedings}.

\bibitem{acar2021recent}
Acar, P., 2021.
\newblock ``Recent progress of uncertainty quantification in small-scale
  materials science''.
\newblock {\em Progress in Materials Science, {\bf 117}}, p.~100723.

\end{thebibliography}



\end{document}